\documentclass[journal]{IEEEtran}
\usepackage{amsmath,amsfonts}
\usepackage{algorithm}
\usepackage{array}
\usepackage[caption=false,font=normalsize,labelfont=sf,textfont=sf]{subfig}
\usepackage{textcomp}
\usepackage{stfloats}
\usepackage{url}
\usepackage{verbatim}
\usepackage{graphicx}
\usepackage{cite}
\usepackage{xcolor}
\usepackage{amssymb}
\usepackage{threeparttable}
\usepackage{algpseudocode}
\usepackage{color}
\usepackage[colorlinks=true, allcolors=blue, bookmarks=true]{hyperref}
\newcommand{\appendixprefix}{}

\renewcommand{\theHequation}{\appendixprefix\arabic{equation}}

\begin{document}

\title{Deep Probabilistic Indoor Gas Source Localization via Physical Dependency-Guided Sequential Inference}

\author{Seunghwan Kim$^{1}$, Hyungjin Kim$^{2}$, Junhee Lee$^{2}$ and Hyondong Oh$^{2}$

\thanks{$^{1}$S. Kim is with the Department of Mechanical Engineering, Ulsan National Institute of Science and Technology (UNIST), Ulsan 44919, Republic of Korea. 
{\tt\footnotesize \{kevin6960\}@unist.ac.kr}
}

\thanks{$^{2}$H. Kim, J. Lee and H. Oh are with the Department of Mechanical Engineering, Korea Advanced Institute of Science and Technology (KAIST), Daejeon 34141, Republic of Korea. 
(Corresponding author: Hyondong Oh)
{\tt\footnotesize \{gudwls124z, ljh0124, h.oh\}@kaist.ac.kr}
}

\thanks{This work has been submitted to the IEEE for possible publication.
Copyright may be transferred without notice, after which this version
may no longer be accessible.}
}

\markboth{}
{}

\maketitle

\begin{abstract}
Reliable gas source localization (GSL) is critical to safety in industrial and urban environments, yet remains challenging indoors because walls and obstacles interact with airflow to create complex gas dispersion.
High-fidelity models such as computational fluid dynamics and filament models can capture these effects, but their computational cost limits online use. 
We propose a deep probabilistic framework that infers the source posterior from sparse and noisy measurements collected by a mobile robot. 
Unlike end-to-end models that directly infer source estimates from measurements, the proposed method incorporates physical dependencies of indoor gas transport, where wind and source location govern the concentration field. 
These dependencies are embedded through sequential conditional inference, in which inferred wind and concentration fields guide source posterior estimation. 
This structure improves localization under sparse and noisy observations. 
Evaluations show that the proposed method outperforms representative GSL baselines and enables accurate and efficient active GSL in simulations. 
Real-robot experiments demonstrate the feasibility of online operation on an embedded GPU.
\end{abstract}


\section{Introduction} \label{sec1}
Accidental gas leaks in industrial and urban environments pose severe risks to human safety, property, and the environment.
Rapid detection and localization of the leak source is therefore critical to minimizing casualties and economic loss.
Traditional monitoring strategies remain limited in practice: fixed sensor networks offer restricted coverage and poor adaptability to changing environments, and manual inspection is both time-consuming and hazardous.
This highlights the need for rapid, safe, and automated solutions.
As a result, there has been growing interest in gas source localization (GSL) using mobile sensing platforms, which can provide wider coverage, more flexible deployment, and faster response than static approaches.

Beyond simply carrying sensors through the environment, mobile robots can actively decide where to collect new gas and wind measurements based on the information gathered so far.
In mobile-robot-based active GSL, the robot sequentially updates the posterior over the source location from sensor measurements and uses this posterior to quantify source-location uncertainty and guide subsequent sensing actions.
By integrating adaptive sensing, probabilistic source inference, and motion planning, mobile-robot-based active GSL can reduce the time, travel distance, and operational cost required for source localization\cite{bourne2019coordinated,bourne2020decentralized,arain2021sniffing,wang2024exploration,Hutchinson17}.
One important component of this process is estimating the source posterior from the measurements collected so far, which requires an observation model that evaluates the likelihood of the measurements under possible source hypotheses.
In GSL, this observation model describes how gas concentration measurements are generated from a hypothesized source and therefore corresponds to the gas dispersion model.

In outdoor scenarios, GSL often relies on simple analytical gas dispersion models, such as the Gaussian dispersion model~\cite{hutchinson2019source} or the isotropic dispersion model~\cite{vergassola2007infotaxis}.
These relatively simple dispersion assumptions have enabled much of the outdoor GSL research to focus primarily on devising efficient search strategies~\cite{hutchinson2019source,vergassola2007infotaxis,hutchinson2018entrotaxis,hutchinson2018information,park2020cooperative,park2022receding,an2022receding}.
In contrast, indoor environments are characterized by walls, obstacles, and their interactions with airflow, which produce complex gas-dispersion patterns and render such simple analytical models inadequate.
Although computational fluid dynamics (CFD) simulations can capture these complex dispersion patterns, their high computational cost makes them unsuitable for online deployment.
Therefore, a central challenge in indoor GSL is to develop source estimation methods that can account for complex indoor gas behavior while remaining computationally feasible for mobile-robot deployment.

Recent studies have begun to address this challenge by introducing more practical dispersion models for indoor robotic GSL.
One important enabling development is the use of filament dispersion models, which are computationally much lighter than CFD while capturing indoor dispersion characteristics more faithfully than prior simplified analytical models, as demonstrated by Ojeda et al.~\cite{ojeda2024robotic}.
However, even filament dispersion models remain computationally demanding for online inference, since they require propagating a large number of stochastic filaments through the airflow at every inference step and across multiple source hypotheses.
As reported in~\cite{ojeda2024robotic}, a single update of the source posterior can take several seconds, which poses a serious limitation for real-time deployment.
This computational bottleneck motivates the exploration of deep learning as a promising alternative, offering strong representational capacity together with fast GPU-based inference~\cite{ruiz2024gas,jin2023towards,zhong2024awed,tian2025deep,nam2026corrected}.

Specifically, we develop a deep learning--based probabilistic GSL framework that infers the source posterior from sparse and noisy measurements of gas concentration and wind collected by a mobile robot in a known indoor layout.
A natural deep learning strategy for indoor GSL is to directly predict the source location from gas and wind measurements in an end-to-end manner~\cite{tian2025deep}.
While conceptually simple, such an end-to-end model does not explicitly exploit the physical process of indoor gas dispersion: gas is released from the source, transported by the wind, and observed as concentrations at the sensors.
As a result, end-to-end models should be able to implicitly recover the underlying dispersion structure from data, which can be difficult under sparse and noisy measurements.

To explicitly exploit the physical dependency structure of indoor gas dispersion, we propose a physical dependency-guided sequential inference framework for GSL.
Instead of directly estimating the source posterior with a single end-to-end model, this approach represents the underlying physical dependencies as successive conditional inference steps, thereby decomposing source inference into a physically guided sequence.
In this sequence, intermediate estimates of the wind and concentration fields guide the final estimation of the source posterior.

By incorporating this physical dependency structure into a deep probabilistic GSL framework, the proposed method enables reliable source inference under sparse and noisy measurements in complex indoor environments.
Furthermore, the wind and concentration fields estimated within this sequential inference process naturally extend the method beyond GSL.
In particular, estimating the spatial distribution of gas concentration corresponds to gas distribution mapping (GDM), which provides information about gas contamination across the environment.
The uncertainty estimates of the wind and concentration fields, together with the source posterior, are further incorporated into an uncertainty-aware utility function for active GSL.

The proposed method is trained on CFD-generated data with noisy sensor simulation to approximate real-world sensing conditions.
We first evaluate its GSL and GDM performance in simulation under varying numbers of randomly sampled observations, thereby assessing its robustness to different levels of measurement sparsity.
The proposed method is then compared with representative indoor GSL and GDM methods to verify its effectiveness in both source inference and concentration-field estimation.
We further assess its autonomous active GSL capability with a mobile robot in simulated environments and demonstrate online embedded-GPU deployment through controlled real-robot experiments in complex multi-room indoor environments.

Overall, the proposed method addresses the online computational burden of high-fidelity dispersion-model-based indoor GSL while extending learning-based source inference to complex multi-room indoor environments.
The contributions of this study are summarized as follows.
\begin{itemize}
    \item We propose, to the best of our knowledge, the first deep probabilistic GSL method for online mobile-robot deployment in complex multi-room indoor environments;

    \item We introduce physical dependency-guided sequential inference within a deep probabilistic GSL framework, allowing the network to exploit the physical dependency structure of indoor gas dispersion for reliable source inference;
     
    \item We provide a unified probabilistic framework that jointly estimates the source posterior and uncertainty-aware wind and concentration field distributions in complex indoor environments, with the concentration-field estimates also supporting GDM; and

    \item We validate the proposed method through GSL and GDM comparisons, autonomous active GSL simulations, and real-robot experiments demonstrating online embedded-GPU deployment in complex multi-room indoor environments.
\end{itemize}
The code, dataset, and experiment videos will be made publicly available upon acceptance.

\section{Related Work} \label{sec:related_work}
This section reviews existing approaches to indoor gas source localization (GSL) and gas distribution mapping (GDM), with an emphasis on their relevance to probabilistic source estimation in complex indoor environments. 
Section~\ref{sec2:gsl} discusses model-based and learning-based GSL methods, while Section~\ref{sec2:gdm} reviews representative GDM approaches, including kernel distribution mapping (KDM), Gaussian Markov random field (GMRF), and recent deep learning--based models.

\subsection{Indoor gas source localization}\label{sec2:gsl}
To handle complex indoor dispersion, early studies approximated the observation likelihood between gas measurements and source locations using kernel density estimation and incorporated it into Bayesian inference frameworks~\cite{prabowo2020bayesian,prabowo2023integration}.
For active GSL, these studies employed either information-based utility planning~\cite{prabowo2020bayesian} or a reactive strategy based on Anemotaxis~\cite{prabowo2023integration}.
However, their reliance on prior simulations to construct environment-specific likelihoods limits their generalizability across indoor layouts.

Other studies introduced simplified analytical or heuristic observation models for indoor GSL.
An indoor Gaussian dispersion model based on obstacle-aware diffusion distance was utilized with a particle filter and a dual-mode planner balancing local search and information-driven global exploration for active GSL~\cite{kim2024gas}.
However, its applicability is largely restricted to low-airflow conditions.
An upstream-biased observation model using local airflow and obstacle information was also combined with Bayesian estimation and information-theoretic search~\cite{ojeda2021information}.
However, such simplified observation assumptions can limit the reliability of source estimation under complex indoor airflow and gas dispersion patterns.

Other studies adopted the filament dispersion model~\cite{farrell2003plume} as a lightweight alternative to CFD, representing turbulent gas transport as stochastic filaments driven by local airflow and diffusion. 
Using GMRF-based wind-field estimation~\cite{monroy2017online}, Ojeda et al.~\cite{ojeda2024robotic} incorporated the filament model into a probabilistic source-estimation framework.
For active GSL, the next sensing location was greedily selected by maximizing an expected information value derived from variations in simulated plume predictions across different source hypotheses~\cite{ojeda2024robotic}. 
Although the filament model provides improved physical fidelity, estimating the source posterior requires repeated simulations across multiple source hypotheses, creating a substantial computational bottleneck for online inference.

To reduce the computational cost of physics-based dispersion models, learning-based surrogate models have been explored for fast gas-dispersion prediction. 
A physics-regularized model was proposed to generate physically plausible concentration fields, even though it was limited to obstacle-free environments~\cite{ruiz2024gas}. 
A U-Net--based surrogate was also combined with Markov chain Monte Carlo and information-theoretic planning for probabilistic active GSL~\cite{jin2023towards}. 
However, this approach was evaluated only in simplified environments with limited obstacle configurations and fixed inlet conditions. 
More generally, although surrogate models accelerate individual dispersion predictions, probabilistic source estimation still requires repeated evaluations over numerous source hypotheses, which can remain computationally burdensome for online active GSL.

To avoid repeated dispersion-model evaluations, recent studies have directly inferred source locations from sparse observations using end-to-end deep learning~\cite{zhong2024awed,tian2025deep}. 
A joint GDM and GSL framework was proposed in~\cite{zhong2024awed}, but it was developed for obstacle-free environments and evaluated under limited sensing trajectories. 
Another end-to-end model was trained in a simple single-room environment using a small number of experimental trials~\cite{tian2025deep}. 
These methods reduce inference cost, but their generalization to complex indoor layouts remains insufficiently validated, and they do not explicitly provide the probabilistic representations required for uncertainty-aware active GSL.

In learning-based active source search, reinforcement learning has also been used to learn search policies for indoor GSL~\cite{chen2021deep,he2024gas}. 
DQN-based approaches formulated source search as a sequential decision-making problem using historical measurements and map information~\cite{chen2021deep} and were later extended to active source search with an olfactory quadruped robot using a dueling DQN~\cite{he2024gas}. 
These studies suggest that reinforcement learning can improve search efficiency, but they do not explicitly maintain a posterior distribution over the source location.
It is worth while noting that reliable probabilistic source estimation plays an important role for updating source-location uncertainty during search and determining when localization is sufficiently confident.
RL-based source search has been combined with explicit posterior estimation over source parameters~\cite{park2022source,lee2025enhanced}; however, these approaches have mainly been developed for outdoor scenarios and are difficult to transfer directly indoors, where walls and obstacles disturb airflow and produce complex gas-dispersion patterns.

In summary, traditional indoor GSL methods either rely on simplified observation assumptions or incur high computational costs.
Learning-based surrogate models support probabilistic source estimation but still require repeated evaluations over many source hypotheses. 
End-to-end methods avoid such repeated dispersion-model evaluations but have mainly been validated in simplified environments and have not been fully extended to probabilistic active GSL. 
These limitations motivate the deep learning--based GSL framework proposed in this paper, which retains the efficiency of end-to-end learning while incorporating the physical dependency structure of gas transport for reliable inference in complex indoor environments. 
For active GSL, the proposed framework also provides probabilistic outputs that support uncertainty-aware sequential decision making.

\subsection{Indoor gas distribution mapping}\label{sec2:gdm}
GDM aims to construct spatial representations of gas concentration from sparse sensor measurements.
One of the earliest approaches, KDM~\cite{lilienthal2004building}, discretizes the environment into a grid and assigns Gaussian-shaped distributions around each measurement; 
KDM+V~\cite{lilienthal2009statistical} extended this to include uncertainty, KDM+V/W~\cite{reggente2009using} incorporated local wind information via anisotropic kernels, and later extensions combined KDM+V with SLAM to account for obstacles~\cite{kamarudin2018integrating, visvanathan2020improved}.
Although lightweight and well-suited for mobile robots, these methods either ignore wind information or exploit it only in a strictly local manner.

To better incorporate physical structure while retaining probabilistic inference, GMRF-based methods model gas concentration as a Gaussian random field on a grid graph, where local conditional dependencies yield a sparse precision matrix that enables efficient inference and allows physical priors to be incorporated through regularization.
G-GMRF~\cite{g2016time} accounts for obstacles by adapting spatial connectivity using the occupancy map, while GW-GMRF~\cite{gongora2020joint} jointly estimates gas and wind fields through a transport-consistency regularizer.
However, online inference still requires repeatedly solving large linear systems over the entire grid graph, which can be computationally demanding for real-time mobile-robot deployment.

GaBP+~\cite{rhodes2023structurally} replaces the global optimization in GMRFs with local Gaussian belief propagation, enabling approximate inference with much lower computational cost.
However, obtaining an accurate converged solution still requires a substantial number of iterations, and GaBP+ does not directly provide meaningful global uncertainty under limited iterations, which limits its utility for active sensing.

To achieve both online efficiency and accurate gas distribution estimates, there is a growing need for deep learning--based approaches.
A convolutional encoder--decoder network for GDM super-resolution was proposed in~\cite{winkler2022super}, followed by RABI-GNN, a bidirectional graph neural network for GDM~\cite{winkler2024gas}.
However, these methods mainly focus on deterministic concentration-field prediction in obstacle-free environments and do not explicitly provide probabilistic representations of gas concentration fields.
This highlights the need for deep probabilistic GDM methods that are computationally efficient and capable of representing uncertainty in complex indoor environments.
Following this direction, the proposed framework in this study treats gas distribution estimation as a structured intermediate component of probabilistic source inference, while still providing fast probabilistic concentration-field estimates.

\section{Deep Probabilistic Gas Source Estimation} \label{sec3}
\begin{figure*}
    \centering                              
    \includegraphics[width=0.75\textwidth]{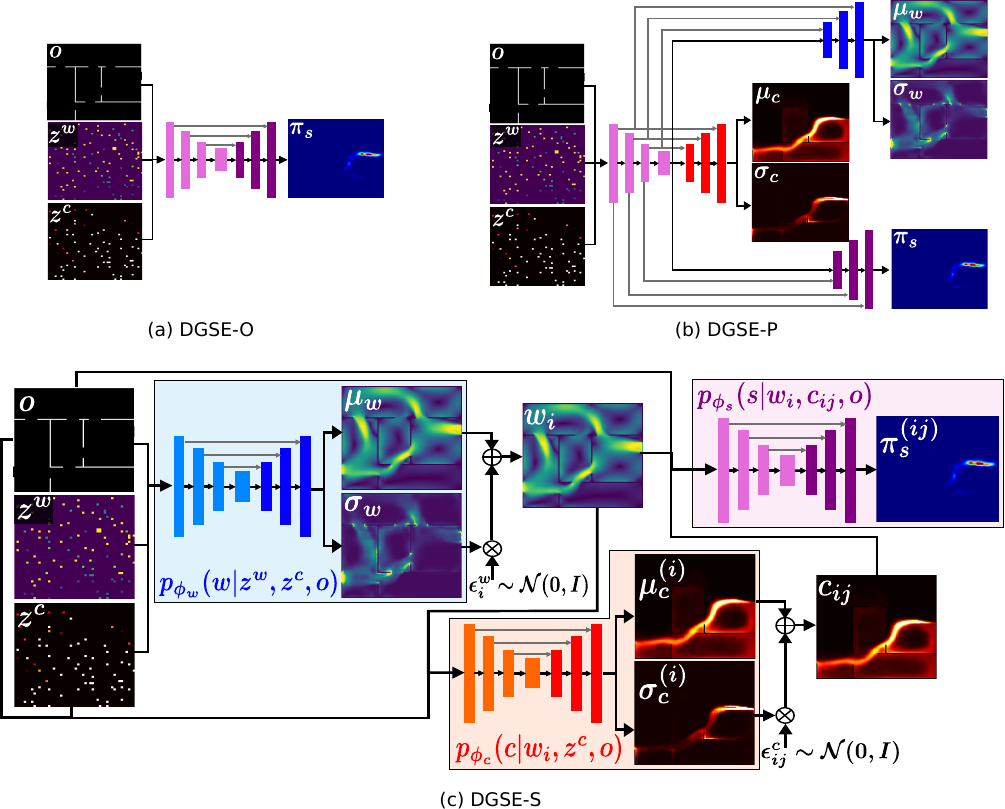}      
    \vspace{-1mm}   
    
    \caption{Network architectures for deep probabilistic gas source estimation (DGSE). 
    (a) DGSE-O directly predicts the source posterior from sparse wind and gas observations and the map. 
    (b) DGSE-P shares an encoder across wind-field (WF), concentration-field (CF), and source-localization (SL) decoders. 
    (c) DGSE-S implements physical dependency-guided sequential inference by conditioning CF estimation on the inferred WF and SL estimation on the inferred WF and CF. } 
    \vspace{-3mm}
    \label{fig1}                                                           
\end{figure*}

\subsection{Problem Formulation and End-to-End Baseline}
Given the limitations of existing approaches discussed earlier, we seek to design a deep learning--based gas source estimation (DGSE) method that can operate effectively in complex indoor environments.
In this work, we consider single-source, steady-state indoor gas dispersion in a known environment map.
The problem of training a neural network to predict the gas source location $s$ based on wind measurements $z^w$, gas concentration measurements $z^c$, and map $o$ can be formulated as
\begin{equation}
\theta^*=\underset{\theta} {\arg\max} \log p_\theta (s_{\text{true}}|z^w, z^c, o),
\end{equation}
where $\theta$ denotes the parameters of the neural network. 
For notational simplicity, all training objectives in this section are expressed for a single sample, with the averaging over training samples omitted.

To apply convolutional architectures, the environment is discretized into $N$ grid cells, and the source location is represented as a categorical variable over these cells. 
The network thus outputs a probability mass function over all grid cells whose entries sum to one, turning gas source localization into a discrete spatial classification problem. 
In this setting, a U-Net~\cite{ronneberger2015u} architecture is a natural choice, as it is widely used in related dense prediction problems.

Before introducing the proposed method, we define a baseline that estimates only the source location from measurements.
This baseline predicts the source posterior directly from the sparse wind measurements, gas concentration measurements, and map.
We denote this source-only estimator as DGSE-O, and its network structure is illustrated in Fig.~\ref{fig1}(a).
DGSE-O is conceptually similar to the method of Tian et al.~\cite{tian2025deep}, as both take gas and wind sensor measurements together with the environmental map as inputs and directly output the source location using a comparable network design.

The output of DGSE-O is modeled as a categorical distribution over all map cells:
\begin{align}
&p_{\theta}(s \mid z^w,z^c,o) = \mathrm{Cat}(s \mid \pi_\theta), \nonumber\\
&\pi_\theta = \mathrm{softmax}(f_\theta(z^w,z^c,o)),
\label{DGSE-O-1}
\end{align}
where $f_\theta(z^w,z^c,o)$ denotes the neural network that produces the logits for the categorical distribution.
The model is trained by maximizing the following likelihood (equivalently, minimizing cross-entropy loss):
\begin{equation}
\log p_{\theta}(s_{\text{true}} \mid z^w,z^c,o) = \sum_{i=1}^N s_{\text{true},i} \log \pi_{\theta,i}, \label{DGSE-O-2}
\end{equation}
where $i$ denotes the $i$-th cell in the discretized environment and $s_{\text{true}} \in \{0,1\}^N$ is a one-hot vector indicating the true source cell.

Although DGSE-O directly estimates the source posterior from sparse observations, it does not explicitly estimate the wind and concentration fields associated with indoor gas dispersion.
We therefore consider two probabilistic extensions of DGSE-O that additionally estimate these fields together with the source posterior.
The two extensions differ in how they incorporate the physical dependencies among source location, wind, and concentration into source inference.

The first extension, DGSE-P, adopts parallel multi-task inference.
A shared encoder and task-specific decoders estimate the wind field, concentration field, and source posterior in parallel.
In DGSE-P, wind- and concentration-field estimation serve as auxiliary tasks, and the physical dependencies among the three quantities are captured only implicitly through the shared latent representation.

The second extension, DGSE-S, implements the proposed physical dependency-guided sequential inference.
In contrast to DGSE-P, DGSE-S explicitly incorporates physical dependencies by decomposing source inference into successive conditional inference steps involving the intermediate wind and concentration fields.
The following subsections describe DGSE-P and DGSE-S in detail.

\subsection{Parallel multi-task inference}
In DGSE-P, parallel multi-task inference is implemented using a shared encoder and task-specific decoders.
The shared encoder extracts a latent representation from the input, while the task-specific decoders generate probabilistic outputs for the wind field, concentration field, and source posterior.
Through the shared latent representation, DGSE-P can capture relationships among these tasks implicitly while maintaining task-specific output distributions.

Given an input observation $\bar{z} \triangleq (z^w, z^c, o)$, 
the shared encoder $f_{\theta_s}$ produces a latent representation $h = f_{\theta_s}(\bar{z})$.
Three task-specific decoders are then applied to $h$.
First, the wind-field decoder parameterized by $\psi_w$ outputs the mean and variance of a Gaussian distribution with diagonal covariance:
\begin{equation}
p_{\theta_s, \psi_w}(w \mid \bar{z}) = \mathcal{N}(w \mid \mu_w(h), \operatorname{diag}(\sigma_w(h)^2)), \label{eq4}
\end{equation}
where $\mu_w(h)$ and $\sigma_w(h)^2$ are the mean and variance predicted by the wind-field decoder.
Second, the concentration field decoder parameterized by $\psi_c$ is defined analogously:
\begin{equation}
p_{\theta_s, \psi_c}(c \mid \bar{z}) = \mathcal{N}(c \mid \mu_c(h), \operatorname{diag}(\sigma_c(h)^2)), \label{eq5}
\end{equation}
where $\mu_c(h)$ and $\sigma_c(h)^2$ are the concentration field decoder outputs.
For computational tractability in online deployment, we model $p_{\theta_s, \psi_w}(w \mid \bar{z})$ and $p_{\theta_s, \psi_c}(c \mid \bar{z})$ using diagonal Gaussians, namely independent per-cell uncertainties. 
While this ignores explicit spatial correlations, it yields a lightweight and scalable predictor that still provides meaningful uncertainty estimates.
Finally, the source-location decoder parameterized by $\psi_s$ produces a categorical distribution:
\begin{equation}
p_{\theta_s, \psi_s}(s \mid \bar{z}) = \mathrm{Cat}(s \mid \pi_s(h)), \label{eq6}
\end{equation}
where $\pi_s(h)=\text{softmax}(f_{\psi_s}(h))$ denotes the class probabilities over the discretized spatial domain.

The total log-likelihood objective is formulated as:
\begin{multline}
\mathcal{J}(\theta_s, \psi_w, \psi_c, \psi_s) =  \lambda_w \log p_{\theta_s, \psi_w}(w_{\text{true}} \mid \bar{z}) \\ + \lambda_c \log p_{\theta_s, \psi_c}(c_{\text{true}} \mid \bar{z}) + \lambda_s \log p_{\theta_s, \psi_s}(s_{\text{true}} \mid \bar{z}), \label{eq7}
\end{multline}
where $\lambda_w, \lambda_c$ and $\lambda_s \geq 0$ are non-negative task weights.
The effect of these weights on training is examined in Appendix~\ref{app:sloss}.
For each task, the corresponding log-likelihood terms in Eq.~\eqref{eq7} are defined as follows.
The log-likelihood of the wind field can be expressed as:
\begin{multline}
\log p_{\theta_s, \psi_w}(w_{\text{true}} \mid \bar{z}) = \\-\frac{1}{2} \sum_{i=1}^N \sum_{k=1}^{2} \left[ \log(2\pi \sigma_{w,ik}^2) + \frac{\left(w_{\text{true},ik} - \mu_{w,ik}\right)^2}{\sigma_{w,ik}^2} \right], \label{eq8}
\end{multline}
where $k \in \{1,2\}$ denotes the components of the 2D wind vector and $i$ indexes the $i$-th cell in the environment discretized into $N$ grid cells.
Here, the predicted variance term $\sigma_w^2$ represents heteroscedastic aleatoric uncertainty~\cite{endall2017uncertainties}, capturing the ambiguity in wind-field reconstruction caused by sparse and noisy sensor observations.

For numerical stability, Eq.~\eqref{eq8} can be reformulated using the log-variance $s_w=\log \sigma_w^2$, expressed as:
 \begin{align}
\log p_{\theta_s, \psi_w}(w_{\text{true}} &\mid \bar{z}) = \nonumber\\  
-\frac{1}{2} \sum_{i=1}^N \sum_{k=1}^{2} [&\log(2\pi)+s_{w,ik} \nonumber\\
&+\exp(-s_{w,ik}) \cdot (w_{\text{true},ik}-\mu_{w,ik})^2 ]. \label{eq9}
\end{align}
Similarly, the concentration field log-likelihood is expressed as:
\begin{align}
\log p_{\theta_s, \psi_c}(c_{\text{true}} \mid \bar{z}) =& \nonumber \\
-\frac{1}{2} \sum_{i=1}^N [\log&(2\pi) +s_{c,i}  \nonumber \\
&+ \exp(-s_{c,i}) \cdot (c_{\text{true},i}-\mu_{c,i})^2], \label{eq10}
\end{align}
where $s_{c,i}=\log \sigma_{c,i}^2$.
Although Eqs.~\eqref{eq9} and~\eqref{eq10} express the wind-field and concentration-field log-likelihoods as sums over spatial cells, the corresponding losses are spatially averaged before applying the task weights in implementation.
This normalization keeps the task-loss scales comparable and prevents larger environments from having disproportionate influence during training.

The source location log-likelihood is represented as:
\begin{equation}
\log p_{\theta_s, \psi_s}(s_{\text{true}} \mid \bar{z}) = \sum_{i=1}^N s_{\text{true},i} \log \pi_{s,i}, \label{eq11}
\end{equation}
where $s_{\text{true}} \in \{0,1\}^N$ is a one-hot vector indicating the true source cell.
We employ a U-Net backbone in which the encoder and each decoder are connected through skip connections, as illustrated in Fig.~\ref{fig1}(b).

\subsection{Physical dependency-guided sequential inference}
\begin{figure}
    \centering                              
    \includegraphics[width=0.2\textwidth]{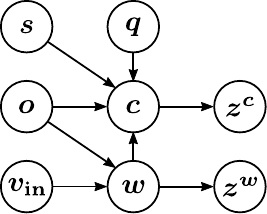}      
    \vspace{-1mm}   
    
    \caption{Bayesian network representing the physical dependencies used to derive the conditional inference sequence in DGSE-S.}  
    \vspace{-4mm}
    \label{fig2}                                                           
\end{figure}
DGSE-S decomposes source inference according to the physical dependency structure of indoor gas dispersion, as represented by the Bayesian network in Fig.~\ref{fig2}.
Specifically, in the indoor gas dispersion process, the wind field is governed by the map $o$ and inlet velocity $v_{\mathrm{in}}$; the concentration field is governed by the source location $s$, source release rate $q$, map $o$, and wind field $w$; and the wind and gas measurements are generated from the corresponding fields.
The inlet velocity $v_{\mathrm{in}}$ and source release rate $q$ are not assumed to be available to the robot and are therefore not provided as explicit network inputs.
Their effects are instead accounted for implicitly through the sparse wind and gas concentration measurements together with the map.
This dependency structure leads to the following two conditional independence relations used in the physical dependency-guided sequential inference.

First, given the wind field, gas measurements, and map, the wind measurements provide no additional information for reconstructing the concentration field;
thus, the following relation holds:
\begin{equation}
p(c \mid w,z^w,z^c,o)=p(c \mid w,z^c,o).\label{cindep}
\end{equation}
Second, once the wind field, concentration field, and map are given, the sensor measurements provide no additional information for estimating the source location,
leading to:
\begin{equation}
p(s \mid w,c,z^w,z^c,o) = p(s \mid w,c,o).\label{sindep}
\end{equation}
Using the first relation, the marginal predictive distribution of the concentration field can be expressed as:
\begin{align}
 p(c|z^w,z^c,o) &= \int p(w|z^w,z^c,o) p(c|w,z^w,z^c,o) dw \nonumber \\
&=\int p(w|z^w,z^c,o) p(c|w,z^c,o) dw \nonumber \\
 &=  \mathbb{E}_{w^* \sim p(w|z^w,z^c,o)} \left[ p(c|w^*,z^c,o)\right]. \label{c_integ}
\end{align}
That is, the concentration field is obtained by marginalizing over the latent wind field.
Similarly, using both relations (Eqs.~\eqref{cindep} and \eqref{sindep}), the marginal source posterior can be expressed as:
\begin{align}
& p(s|z^w,z^c,o) \nonumber \\ 
 &= \int \int p(w,c|z^w,z^c,o) p(s|w,c,z^w,z^c,o) dc\,dw  \nonumber \\
 &= \int \int p(w,c|z^w,z^c,o) p(s|w,c,o) dc\,dw \nonumber \\
 &= \int \int p(w|z^w,z^c,o) p(c|w,z^w,z^c,o) p(s|w,c,o) dc\,dw  \nonumber \\
 &= \int \int p(w|z^w,z^c,o) p(c|w,z^c,o) p(s|w,c,o) dc\,dw  \nonumber\\
 &= \int p(w|z^w,z^c,o) \int  p(c|w,z^c,o) p(s|w,c,o) dc\,dw  \nonumber \\ 
 &=  \mathbb{E}_{ w^* \sim p(w|z^w,z^c,o)} \left[ \mathbb{E}_{ c^* \sim p(c|w^*,z^c,o)} \left[ p(s|w^*,c^*,o)\right] \right].\label{s_integ}
\end{align}
That is, the source posterior is obtained by marginalizing sequentially over the latent wind and concentration fields.

If the conditional distributions $p(w|z^w,z^c,o)$, $p(c|w,z^c,o)$, and $p(s|w,c,o)$ are approximated by separate subnetworks parameterized by $\phi_w$, $\phi_c$, and $\phi_s$, respectively, each subnetwork follows the output distribution used in Eqs.~\eqref{eq4}--\eqref{eq6}.
Specifically, $p_{\phi_w}$ outputs $(\mu_w,\sigma_w)$ for a diagonal Gaussian distribution over $w$, $p_{\phi_c}$ outputs $(\mu_c,\sigma_c)$ for a diagonal Gaussian distribution over $c$, and $p_{\phi_s}$ outputs the categorical distribution $\pi_s$ as illustrated in Fig.~\ref{fig1}(c).

With these conditional predictors, the corresponding marginal log-likelihood objective can be expressed as:
\begin{align}
 \mathcal{J}(&\phi_w, \phi_c, \phi_s) = \lambda_w \log p_{\phi_w}(w_{\text{true}}|z^w,z^c,o) \nonumber \\
  & + \lambda_c \log \mathbb{E}_{w^* \sim p_{\phi_w}} \left[ p_{\phi_c}(c_{\text{true}}|w^*,z^c,o)\right] \nonumber \\
  &+ \lambda_s \log \mathbb{E}_{ w^* \sim p_{\phi_w}} \left[ \mathbb{E}_{ c^* \sim p_{\phi_c}} \left[p_{\phi_s}(s_{\text{true}}|w^*,c^*,o)\right] \right].
\label{eq14}
\end{align}
For brevity, $w^*\sim p_{\phi_w}$ and $c^*\sim p_{\phi_c}$ denote samples
from the corresponding conditional distributions, with conditioning variables omitted when clear from context.

During training, supervision for $w$, $c$, and $s$ is used to learn the conditional predictors that constitute DGSE-S.
At inference time, $w$ and $c$ serve as intermediate random variables in the physical dependency-guided sequential inference process and are marginalized out to obtain $p(c\mid z^w,z^c,o)$ and $p(s\mid z^w,z^c,o)$ as in Eqs.~\eqref{c_integ} and~\eqref{s_integ}, respectively.
Since these marginalizations are analytically intractable, we approximate the corresponding expectations by Monte Carlo sampling with a finite number of samples.
The predictive distributions of the concentration field and source location are then expressed as:
\begin{align}
p&(c|z^w,z^c,o) = \mathbb{E}_{w^* \sim p_{\phi_w}} \left[ p_{\phi_c}(c|w^*,z^c,o)\right]\nonumber  \\ 
&\approx  \frac{1}{K} \sum_{i=1}^K p_{\phi_c}(c|w_i,z^c,o), \quad w_i \sim p_{\phi_w}(w|z^w,z^c,o),\label{inf_c}
\end{align}
\begin{align}
p(s|z^w,z^c,o) &= \mathbb{E}_{ w^* \sim p_{\phi_w}} \left[ \mathbb{E}_{ c^* \sim p_{\phi_c}} \left[p_{\phi_s}(s|w^*,c^*,o)\right] \right] \nonumber  \\
&\approx \frac{1}{KT} \sum_{i=1}^K \sum_{j=1}^T p_{\phi_s}(s|w_i,c_{ij},o),  \label{inf_s}\\ 
 w_i \sim p_{\phi_w}&(w|z^w,z^c,o), \quad c_{ij} \sim p_{\phi_c}(c|w_i,z^c,o), \nonumber
\end{align}
where $K$ and $T$ are user-controlled parameters that trade accuracy for latency, and the computations across samples are parallel and amenable to GPU batching.
Samples passed between modules are obtained using the reparameterization trick~\cite{kingma2013auto}:
\begin{align}
w_i &= \mu_w + \sigma_w \odot \epsilon_i^w,
\qquad \epsilon_i^w \sim \mathcal{N}(0,I), \label{eq:reparam_w}\\
c_{ij} &= \mu_c^{(i)} + \sigma_c^{(i)} \odot \epsilon_{ij}^c,
\qquad \epsilon_{ij}^c \sim \mathcal{N}(0,I), \label{eq:reparam_c}
\end{align}
where $\odot$ denotes element-wise multiplication, and 
$(\mu_c^{(i)},\sigma_c^{(i)})$ denote the concentration-subnetwork outputs conditioned on the sampled wind field $w_i$.
This sampling process corresponds to the module-to-module inputs shown in Fig.~\ref{fig1}(c).

The marginal predictive distribution of the concentration field obtained via sampling in Eq.~\eqref{inf_c} is a mixture of Gaussians.
For computational simplicity and to obtain a compact probabilistic field map, we approximate this mixture with a single diagonal Gaussian by moment matching.
Specifically, for $K$ samples, we compute 
\begin{align}
\bar{\mu}_c &= \frac{1}{K}\sum_{i=1}^{K}\mu_{c}^{(i)}, \label{inf_c_1} \\
\bar{\sigma}_c^2 &= \frac{1}{K}\sum_{i=1}^{K}\left(\sigma_{c}^{2\,(i)}+\mu_{c}^{(i)2}\right)-\bar{\mu}_c^{\,2}. \label{inf_c_2} 
\end{align}
The marginal predictive distribution of the source location obtained via sampling in Eq.~\eqref{inf_s} is computed as:
\begin{equation}
\bar{\pi}_s
=
\frac{1}{KT}
\sum_{i=1}^{K}
\sum_{j=1}^{T}
\pi_s^{(ij)},
\label{inf_s_prob}
\end{equation}
where $\pi_s^{(ij)}$ denotes the source-probability vector predicted by the source subnetwork conditioned on the sampled wind and concentration fields $(w_i,c_{ij})$.

The same marginalization structure also appears when formulating the training objective in Eq.~\eqref{eq14}.
However, the marginal likelihood terms in Eq.~\eqref{eq14} do not admit closed-form evaluation, and Monte Carlo approximation is therefore required.
Because each expectation appears inside a logarithm, a finite-sample Monte Carlo estimate of the marginal log-likelihood is biased.
Reducing this bias by substantially increasing the number of samples would be computationally impractical in our setting.
To obtain a tractable alternative, we instead apply Jensen's inequality and derive a surrogate objective, namely a lower bound, expressed as:
\begin{align}
 \mathcal{J}(&\phi_w,\phi_c,\phi_s) \ge \mathcal{L}(\phi_w,\phi_c,\phi_s)\nonumber \\
&= \lambda_w \log p_{\phi_w}(w_{\text{true}} \mid z^w, z^c, o) \nonumber \\ 
&+ \lambda_c \mathbb{E}_{w^\ast \sim p_{\phi_w}} 
\left[\log p_{\phi_c}(c_{\text{true}} \mid w^\ast, z^c, o)\right] \nonumber \\
&+ \lambda_s \mathbb{E}_{w^\ast \sim p_{\phi_w}}
\left[
\mathbb{E}_{c^\ast \sim p_{\phi_c}}
\left[\log p_{\phi_s}(s_{\text{true}} \mid w^\ast, c^\ast, o)\right]
\right].
\label{eq15}
\end{align}

The lower bound $\mathcal{L}$ in Eq.~\eqref{eq15} can be approximated with Monte Carlo sampling. 
This approximation yields an unbiased estimator of the lower bound, which can be expressed as:
\begin{align}
\mathcal{L}(\phi_w, \phi_c, \phi_s) &\approx \lambda_w \log p_{\phi_w}(w_{\text{true}}|z^w,z^c,o) \nonumber \\ 
&+  \frac{\lambda_c}{R_w} \sum_{i=1}^{R_w} \log p_{\phi_c}(c_{\text{true}}|w_i,z^c,o) \nonumber \\
 &+\frac{\lambda_s }{R_w R_c} \sum_{j=1}^{R_c} \sum_{i=1}^{R_w}\log p_{\phi_s}(s_{\text{true}}|w_i,c_{ij},o),   \label{eq16}
 \end{align}
where $w_i \sim p_{\phi_w}(w \mid z^w, z^c, o)$ and $c_{ij} \sim p_{\phi_c}(c \mid w_i, z^c, o)$.
In practice, following a common approach to reparameterized stochastic training~\cite{kingma2013auto}, we use a single-sample estimator ($R_w=R_c=1$) together with a batch size of 300.
Under the reparameterization trick, this yields a stochastic gradient estimator of the lower bound. 
Empirically, increasing $R_w$ and $R_c$ yielded negligible improvements at higher computational cost.

The log-likelihood terms $\log p_{\phi_w}$, $\log p_{\phi_c}$, and $\log p_{\phi_s}$ appearing in Eq.~\eqref{eq16} can be computed in the same manner as in Eqs.~\eqref{eq9},~\eqref{eq10}, and~\eqref{eq11}, respectively. 
Each conditional predictor in DGSE-S is implemented as a U-Net, and the resulting physical dependency-guided sequential inference architecture is illustrated in Fig.~\ref{fig1}(c).

Although tighter marginal-likelihood bounds, such as IWAE~\cite{burda2015importance}, could be adopted, we found the simple Jensen's inequality--based lower bound sufficient in our setting.
Its empirical consistency with MC estimates of the original marginal objective is analyzed in Appendix~\ref{app:val_object}.

\subsection{Uncertainty-aware utility-based active GSL}\label{sec3_active}
\begin{algorithm}[t]
\caption{Active GSL using DGSE-S probabilistic estimates. $\mathcal{Z}^w$ and $\mathcal{Z}^c$ denote the sets of wind and gas concentration measurements accumulated over time.}
\label{alg:search}
\begin{algorithmic}[1]
\Require Map $o$, probabilistic estimator $\mathcal{E}_{\theta}$, entropy threshold $\eta$
\State $\mathcal{Z}^w \leftarrow \emptyset$, \; $\mathcal{Z}^c \leftarrow \emptyset$
\Repeat
    \State \textit{// Sensing}
    \State $\mathcal{Z}^w \leftarrow \mathcal{Z}^w \cup \{z^w\}$, \quad $\mathcal{Z}^c \leftarrow \mathcal{Z}^c \cup \{z^c\}$

    \State \textit{// Probabilistic estimation}
    \State $\mu_w, \sigma_w^2, \mu_c, \sigma_c^2, \pi_s \leftarrow \mathcal{E}_{\theta}(\mathcal{Z}^w, \mathcal{Z}^c, o)$

    \State \textit{// Termination check}
    \If{$H(\pi_s) \leq \eta \cdot \log N$}
        \State \textbf{break}
    \EndIf

    \State \textit{// Goal selection}
    \State $U_i \leftarrow \exp(-w_d \cdot dist_i)(w_w\sigma^2_{w,i} + w_c\sigma^2_{c,i} + w_s\pi_{s,i})$
    \State Select the next goal cell as $i^\star = \arg\max_{i} U_i$
    \State Move to cell $i^\star$
\Until{time limit exceeded}
\State \Return $\hat{s} = \sum_{(i,j)} \pi_s(i,j) \cdot (r_i, c_j)$
\end{algorithmic}
\end{algorithm}

To examine whether the probabilistic estimates produced by the DGSE variants can support active GSL, 
we integrate them into a simple utility-based planner.
This section does not aim to introduce a new planning algorithm.
Instead, we use a fixed utility function to evaluate whether the estimated source posterior and field uncertainties provide useful guidance for active GSL.

At each search step, the estimator takes the accumulated wind measurements, gas concentration measurements, and map as inputs.
It then produces a source-posterior map $\pi_s$, a wind-field variance map $\sigma_w^2$, and a concentration-field variance map $\sigma_c^2$.
For DGSE-P, these quantities are obtained directly from the parallel multi-task outputs.
For DGSE-S, the source-posterior map and concentration-field variance correspond to $\bar{\pi}_s$ and $\bar{\sigma}_c^2$ in Eqs.~\eqref{inf_s_prob} and~\eqref{inf_c_2}, respectively.
For simplicity, we omit the overbars when referring to these quantities in the common active GSL formulation below.

The next goal cell is selected by maximizing the following utility:
\begin{equation}
U_i = \exp(-w_d \cdot dist_i)
\cdot
\left(
w_w\sigma_{w,i}^2
+
w_c \sigma_{c,i}^2
+
w_s\pi_{s,i}
\right),
\end{equation}
where $i$ denotes the index of the $i$-th cell in an environment discretized into $N$ grid cells, where $N$ is identical to the dimension of the DGSE source posterior $\pi_s$.
The term $\sigma_{w,i}^2 = \sigma_{w,ix}^2 + \sigma_{w,iy}^2$ denotes the aggregated wind-field variance at cell $i$ over the two wind-vector components.
Here, $dist_i$ is the path length from the current robot position to cell $i$, computed for all cells using Dijkstra's algorithm.
The distance term discourages unnecessarily long motions, while the remaining terms encourage the robot to visit regions with high source probability or high uncertainty in the reconstructed physical fields.
The weights $(w_d, w_w, w_c, w_s) = (0.1, 0.25, 2.0, 100.0)$ are fixed across all active GSL environments in Section~\ref{sec_e} and are not re-tuned per environment.

The search terminates when the entropy of the source posterior falls below a user-defined fraction of the maximum entropy:
\begin{equation}
H(\pi_s) \leq \eta \log N,
\end{equation}
where $H(\pi_s) = -\sum_i \pi_{s,i} \log \pi_{s,i}$, $\log N$ is the entropy of the uniform distribution over $N$ grid cells, and $\eta \in [0,1]$ is a user-defined termination threshold.
A smaller value of $\eta$ requires a more concentrated source posterior before termination.

Upon termination, the source location is estimated as the probability-weighted mean:
\begin{equation}
\hat{s} = \sum_{(i,j)} \pi_s(i,j) \cdot (r_i, c_j),
\end{equation}
where $r_i$ and $c_j$ denote the row and column coordinates of cell $(i,j)$ in the discretized grid.

Overall, this active GSL procedure is intentionally kept simple.
Its purpose is to evaluate whether the probabilistic outputs of the DGSE variants are informative enough to support downstream source search decisions, rather than to optimize the search policy itself.
The overall procedure is summarized in Algorithm~\ref{alg:search}.

\section{Data Generation}\label{sec_data}
\begin{table}[t]
\centering
\begin{threeparttable}
\caption{Sensor noise models and parameter ranges.}
\label{tab:noise}
\renewcommand{\arraystretch}{1.2}
\begin{tabular}{lll}
\hline
\textbf{Sensor} & \textbf{Model} & \textbf{Parameters} \\
\hline
Gas conc. & $z_{\text{raw}}^{c} \sim \mathcal{N}(c,\,\sigma_g^2)$ & $\sigma_g = \alpha_g c + \sigma_{env}$ \\
 & & $\alpha_g \in [0,\,1.0]$ \\
 & & $\sigma_{env} \in [0,\,0.5]\times\bar{c}$ \\
\hline
Wind speed & $z^{w,m} \sim \mathcal{N}(w^m,\,\sigma_{w,m}^2)$ & $\sigma_{w,m} = \alpha_w w^m$ \\
 & & $\alpha_w \in [0,\,0.5]$ \\
\hline
Wind dir. & $z^{w,a} \sim \mathcal{N}(w^a,\,\sigma_{w,a}^2)$ & $\sigma_{w,a} \in [0^\circ,\,15^\circ]$ \\
\hline
\end{tabular}
\begin{tablenotes}
\footnotesize
\item[*] $\bar{c}$ denotes the mean concentration of the environment.
\end{tablenotes}
\end{threeparttable}
\vspace{-1mm}
\end{table}

\begin{figure}[t!]
    \centering                              
    \includegraphics[width=0.45\textwidth]{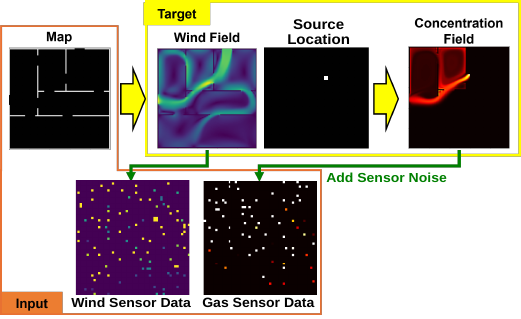}      
    \vspace{-1mm}   
    \caption{Data generation pipeline. Computational fluid dynamics (CFD) simulations generate ground-truth wind fields, concentration fields, and source locations from randomized indoor layouts and boundary conditions. Sparse wind and gas observations are then produced by adding randomized sensor noise before rasterizing the inputs for DGSE training.}   
    \vspace{-3mm}
    \label{fig3}                                                           
\end{figure}

Having defined the model architecture and training objective, 
we now describe the data generation pipeline used to train the proposed method. A key challenge is exposing the model to the full diversity of indoor dispersion conditions, including various layouts, inlet velocities, emission rates, and sensing configurations. 
We address this through CFD-based simulations combined with realistic sensor noise modeling, as summarized in Fig.~\ref{fig3}.

For training data generation, we first created 3,750 indoor layouts. Each layout was defined on a rectangular domain whose width and height were independently sampled from $[5\,\mathrm{m}, 10\,\mathrm{m}]$. The domain was randomly partitioned along vertical and horizontal directions to form multiple rooms separated by walls, with passageways inserted at random locations to ensure connectivity. One flow inlet and one flow outlet were placed at random positions along the outer boundary, and each layout was discretized into a $0.1\,\mathrm{m}$ resolution grid for simulation and learning.

Using these layouts, wind and gas concentration fields were simulated with OpenFOAM~\cite{jasak2009openfoam}. Inlet velocities were randomly selected from $[0.5\,\mathrm{m/s}, 5.0\,\mathrm{m/s}]$. Meshes were generated using \texttt{blockMesh} and \texttt{snappyHexMesh}, and steady-state wind fields were computed with \texttt{simpleFoam} under a RANS turbulence closure (standard $k$--$\epsilon$ model) with velocity-inlet and pressure-outlet boundary conditions. 
For each wind field, four gas dispersion simulations were generated, each with a randomized single source (release rate sampled from $[0.01\,\mathrm{g/s}, 5.0\,\mathrm{g/s}]$ of ethanol), yielding 15,000 environments in total. 
Gas dispersion was computed by solving a passive-scalar advection--diffusion equation using \texttt{scalarTransportFoam}, and the converged steady field was used as the ground-truth concentration. 
The dataset was split at the layout level, with 3,000, 500, and 250 layouts assigned to training, validation, and testing, respectively, yielding 12,000 training, 2,000 validation, and 1,000 test dispersion environments after generating four source configurations per airflow field.

\subsubsection*{Sensor noise simulation}
Sensor noise parameters were randomized for each sample according to the models and ranges summarized in Table~\ref{tab:noise}.
The gas sensor noise ranges are designed to encompass those commonly employed in prior GSL studies using Gaussian sensor models~\cite{hutchinson2018information, an2022receding}, 
while the wind noise ranges cover the specifications of commercial wind sensors such as the TriSonica LI-550.
After noise simulation, the wind data were converted into $x$ and $y$ component vectors for use as an input.

\subsubsection*{Training and input configuration}
For each of the 12,000 training environments, 30 sensor configurations were independently generated on the fly during each epoch.
For each configuration, the number of sensors was sampled uniformly from the discrete set of multiples of five between 5 and 300, namely $\{5,10,15,\ldots,300\}$.
The sensor positions were randomized and independent noise was added, yielding $12{,}000 \times 30 = 360{,}000$ distinct environment--sensing pairs per epoch.
In addition, one of six symmetry augmentations (identity, $90^\circ/180^\circ/270^\circ$ rotations, and $x/y$-reflections) was randomly applied to each pair at load time, so that the effective diversity of samples seen across epochs was further increased without changing the per-epoch sample count. 
For validation and test, 30 sensor configurations per environment were fixed prior to evaluation, yielding 60,000 validation samples and 30,000 test samples.

All observations were rasterized onto the $0.1\,\mathrm{m}$ grid. 
Gas concentration measurements were log-transformed as $z^c = \log(z_{\text{raw}}^{c} + 1)$ after clipping negatives, with unobserved cells set to $-1$.
The ground-truth concentration-field target was transformed using the same logarithmic mapping. 
Wind measurements were represented by two rasterized channels containing their $x$- and $y$-components, with unobserved cells set to zero, while the environment layout was represented by a binary occupancy grid.
To support batch training across variable-size environments, all data were embedded into a fixed $100\times100$ grid canvas with a valid-region mask.
Throughout this paper, $N$ denotes the number of cells in the valid environment region after excluding padded cells, and all losses and evaluation metrics are computed only over this region.
Together, these components form the five-channel input.
The training targets consist of CFD-simulated wind fields, concentration fields, and a one-hot source-location map, all at the same grid resolution. 

We note that training with randomly placed sensors, rather than trajectory-based placements as encountered at deployment, is intentional: it exposes the network to maximally diverse spatial configurations, preventing overfitting to particular trajectory shapes. 
As demonstrated in Sections~\ref{sec_e:indomain} and~\ref{sec_e:real}, models trained in this manner transferred effectively to sequential observations collected during active GSL in our experiments.

\begin{figure*}[t!]
    \centering                              
    \includegraphics[width=0.9\textwidth]{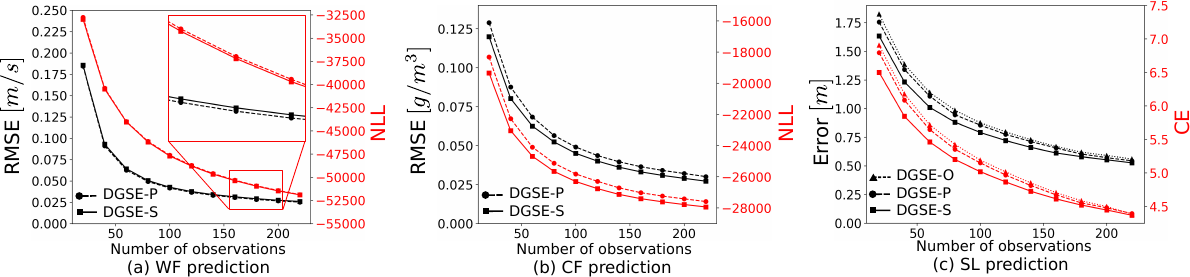}      
    \vspace{-3mm}   
    
    \caption{Quantitative comparison of DGSE-O, DGSE-P, and DGSE-S across 1,000 test environments with varying numbers of observations.
    The three prediction tasks are evaluated: wind-field (WF) prediction, concentration-field (CF) prediction, and source-localization (SL) prediction.
    Black lines denote RMSE for WF/CF and localization error for SL, while red lines denote Gaussian NLL for WF/CF and cross-entropy (CE) for SL.} 
    \vspace{-1mm}
    \label{fig5}                                                           
\end{figure*}

\section{Experimental Validation}\label{sec_e}
This section evaluates the proposed DGSE method from estimator-level performance to autonomous active GSL. 
We first compare the DGSE variants in terms of wind-field (WF) estimation, concentration-field (CF) estimation, and source-localization (SL) performance.
We then compare DGSE-S with representative indoor GDM methods under sparse observations, since the proposed framework jointly supports both GDM and GSL by explicitly estimating the concentration field as an intermediate physical representation.
We further compare DGSE-S with representative indoor GSL methods to evaluate its source-localization performance under the same sparse-observation setting.
This is followed by active GSL simulations and controlled real-robot experiments.

Unless otherwise stated, DGSE-P and DGSE-S were trained with $\lambda_w=\lambda_c=1.0$ and $\lambda_s=0.05$, which were selected based on validation SL loss and training stability. 
During inference, DGSE-S uses $K=T=5$ samples for sample-based marginalization in Eqs.~\eqref{inf_c} and~\eqref{inf_s}.
Detailed analyses of task-weight selection, the surrogate training objective, and the sample-count/latency trade-off are provided in Appendices~\ref{app:sloss}, \ref{app:val_object}, and \ref{app:sample}, respectively. 
The number of parameters, FLOPs, and full training configuration are also
summarized in Appendix~\ref{app:sloss}.

\subsection{Quantitative Comparison of DGSE Variants}\label{sec_e:compare}
We first compare the three DGSE variants---the source-only baseline (DGSE-O), the parallel multi-task model (DGSE-P), and the proposed sequential-inference model (DGSE-S)---in terms of WF, CF, and SL prediction.
For each DGSE variant, we selected the checkpoint at the epoch with the lowest
validation SL loss, using the task-weight setting described above.
The selected checkpoints were evaluated on 1,000 test environments under
varying numbers of observations, as shown in Fig.~\ref{fig5}.
Each test sample includes sensor noise simulated according to the procedure described in Section~\ref{sec_data}.
For WF and CF estimation, performance is measured using root mean square error (RMSE) and Gaussian negative log-likelihood (NLL). 
For SL, performance is measured using source-localization error and cross-entropy (CE).
The source-localization error is computed as the Euclidean distance between the probability-weighted source location estimate $\hat{s}$ defined in Section~\ref{sec3_active} and the ground-truth source location.
The quantitative results are summarized in Fig.~\ref{fig5}.

DGSE-P and DGSE-S show similar WF-prediction performance across the evaluated observation settings.
In CF prediction, DGSE-S achieved lower RMSE and NLL than DGSE-P under the same settings.
For SL, DGSE-S achieved the best overall performance in both localization error and CE.
DGSE-P also improved over DGSE-O, suggesting that intermediate wind field and concentration field prediction provides useful auxiliary supervision for source localization.
The additional improvement of DGSE-S over DGSE-P suggests that explicitly reflecting the physical dependencies underlying indoor gas dispersion can further improve source-localization performance.

\begin{figure*}[t!]
    \centering                              
    % Flattened at 600 DPI to avoid transparency-mask seams in print drivers.
    \includegraphics[width=0.95\textwidth]{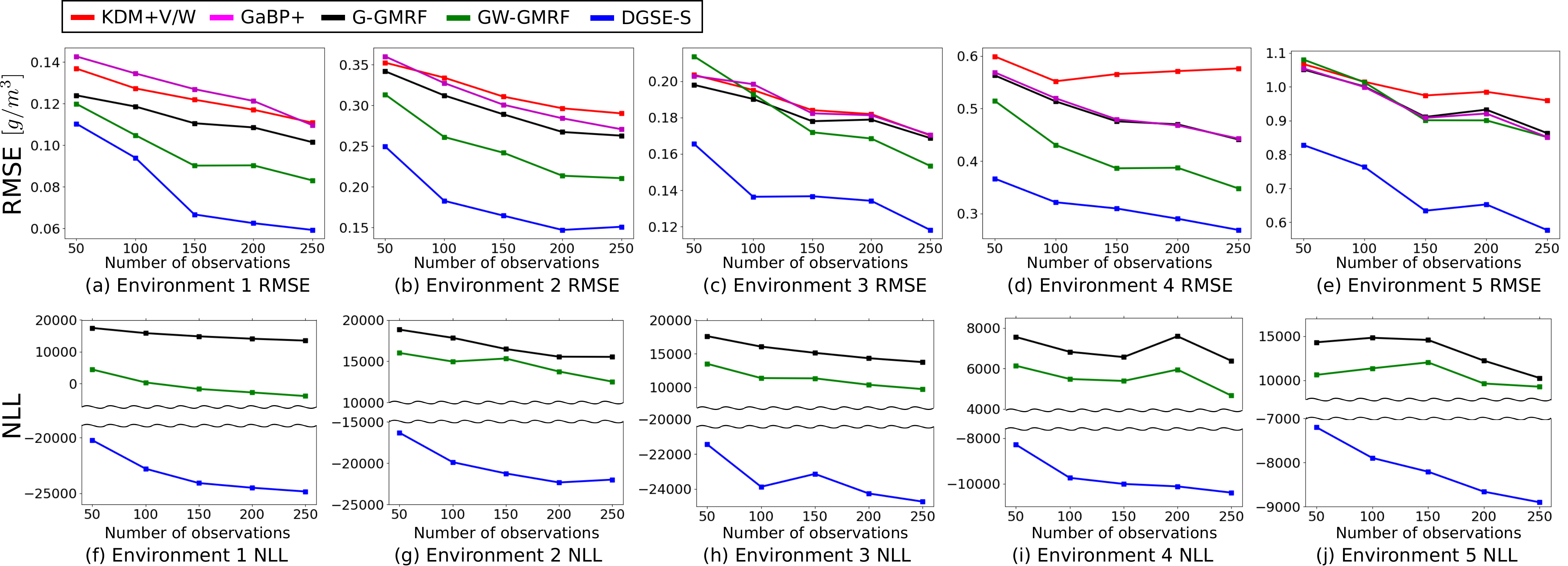}      
    \vspace{-2.5mm}   
    
    \caption{Comparison of DGSE-S and baseline gas distribution mapping (GDM) methods in terms of concentration-field root mean square error (RMSE) and negative log-likelihood (NLL) across five test environments.}   
    \vspace{-2mm}
    \label{fig8}                                                   
\end{figure*}

\begin{figure*}[t!]
    \centering                              
    \includegraphics[width=0.8\textwidth]{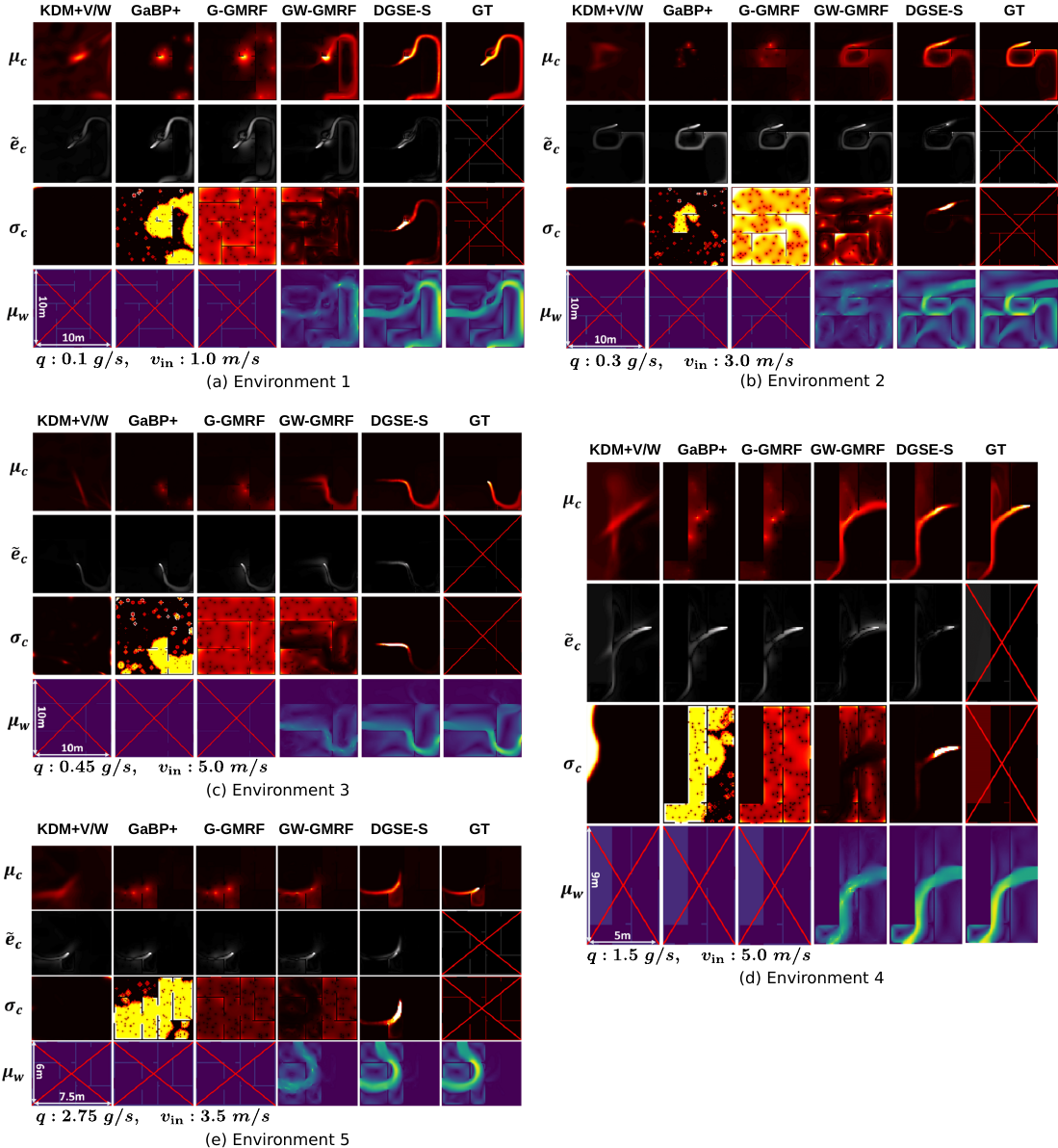}      
    \vspace{-3mm}   
    
    \caption{
    Representative gas distribution mapping (GDM) results for the case of 100 observations across the five test environments, each with a different inlet velocity $v_{\mathrm{in}}$ and release rate $q$.
Each environment compares DGSE-S with KDM+V/W, GaBP+, G-GMRF, and GW-GMRF; GT denotes the ground-truth concentration field.
The predicted concentration-field mean $\mu_c$ is visualized as the estimated gas distribution map, and the predicted wind-field mean $\mu_w$ is shown to indicate the corresponding airflow estimate.
Error maps show $\tilde{e}_c = \|\mu_c - c_{\text{true}}\|$, and uncertainty maps show the predicted standard deviation $\sigma_c$.}
    \vspace{-4mm}
    \label{fig7}                                                           
\end{figure*}

\subsection{Comparison with existing indoor GDM methods}\label{sec_e:gdm_comparison}
\begin{figure}[t!]
    \centering                              
    \includegraphics[width=0.425\textwidth]{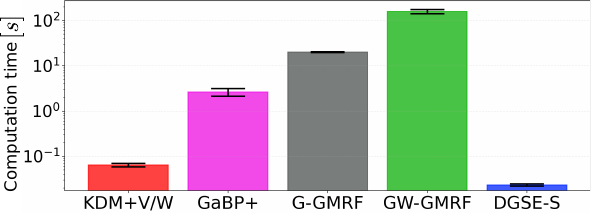}      
    \vspace{-3mm}   
    
    \caption{Computation times of gas distribution mapping (GDM) methods for the case of 100 observations. 
    Times are reported as deployment latencies under the method-specific hardware described in the text.}
    \vspace{-3mm}
    \label{fig9}                                                   
\end{figure}

Building on the above analysis of the DGSE variants, we next compare DGSE-S against traditional indoor GDM approaches on five different test environments.
For each environment, we considered observation numbers of 50, 100, 150, 200, and 250, enabling us to assess performance under varying levels of data sparsity.
At each observation level, 20 independent trials were conducted, where the number of observations was fixed but the sensor locations varied.
Results are reported in terms of the concentration-field RMSE and NLL across trials.

As discussed in Section~\ref{sec2:gdm}, existing learning-based GDM methods have not been demonstrated for complex indoor environments.
We therefore compare DGSE-S with four representative non-learning probabilistic GDM methods applicable to this setting: KDM+V/W~\cite{reggente2009using}, GaBP+~\cite{rhodes2023structurally}, G-GMRF~\cite{g2016time}, and GW-GMRF~\cite{gongora2020joint}.
Note that the variance output of KDM+V/W reflects confidence rather than statistical uncertainty, and GaBP+ does not provide a global probability map; therefore, these methods were excluded from the NLL evaluation.
For comparison with baseline GDM methods, DGSE-S predictions were transformed back to the original scale before computing RMSE and NLL.

The RMSE and NLL as functions of the number of observations are shown in Fig.~\ref{fig8}.
Across all five environments, DGSE-S consistently achieved the lowest RMSE among the evaluated GDM methods and the lowest NLL among the methods providing comparable probabilistic concentration-field outputs.
Representative qualitative results for the case of 100 observations across the five test environments are provided in Fig.~\ref{fig7}, together with the inlet velocities and release rates of each environment.
As shown in Fig.~\ref{fig7}, the prediction error $\tilde{e}_c = \|\mu_c - c_{\text{true}}\|$ and the predicted standard deviation $\sigma_c$ exhibit similar spatial patterns, indicating that the predicted uncertainty reflects regions of high concentration-field reconstruction error.

To assess the computational feasibility of each method for online mobile robot deployment, each method was run on hardware suitable for onboard mobile robot use.
Our method relies on neural network inference and was executed on an NVIDIA Jetson AGX Orin (32 GB), which provides efficient embedded GPU computation suitable for mobile robot deployment.
The comparison methods are classical algorithmic approaches whose computation is primarily CPU-bound; we therefore ran them on an Intel NUC equipped with an Intel Core i7-1360P CPU (12 cores / 16 threads, up to 5.0 GHz), which represents an appropriate high-performance CPU platform for such algorithms.
Accordingly, the reported runtimes should be interpreted as end-to-end deployment latencies under method-appropriate hardware.
The computation times for the case of 100 observations are summarized in Fig.~\ref{fig9}, where DGSE-S exhibited the lowest deployment latency among the compared methods, reflecting the efficiency of embedded-GPU inference.

\subsection{Source localization from sparse random observations}\label{sec_e:gsl_only}
\begin{figure}[t]
    \centering
    \includegraphics[width=0.48\textwidth]{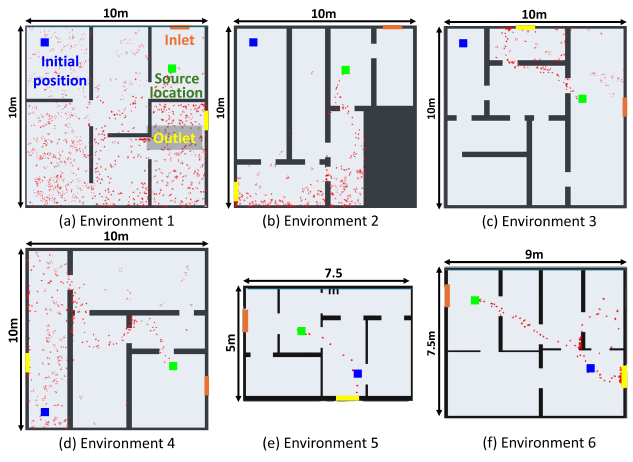}
    \vspace{-3mm}

    \caption{Test environments used for source localization and active GSL evaluations.
    The red dots denote simulated clusters of ethanol gas molecules.}
    \vspace{-1mm}
    \label{fig:gaden_envs}
\end{figure}

Before evaluating active GSL, we first compared the source-localization performance of each estimator under fixed sparse observations.
This experiment isolates the quality of the estimated source posterior from the effects of active goal selection, robot trajectory, and termination criteria.
The evaluation was conducted in the six GADEN~\cite{monroy2017gaden} environments shown in Fig.~\ref{fig:gaden_envs}.
These environments were designed with different indoor layouts, inlet velocities, and ethanol release rates to represent diverse indoor dispersion regimes.
The inlet velocities and release rates of the individual environments are reported in Table~\ref{tab:active_search_results}.

\begin{figure}[t]
    \centering
    \includegraphics[width=0.48\textwidth]{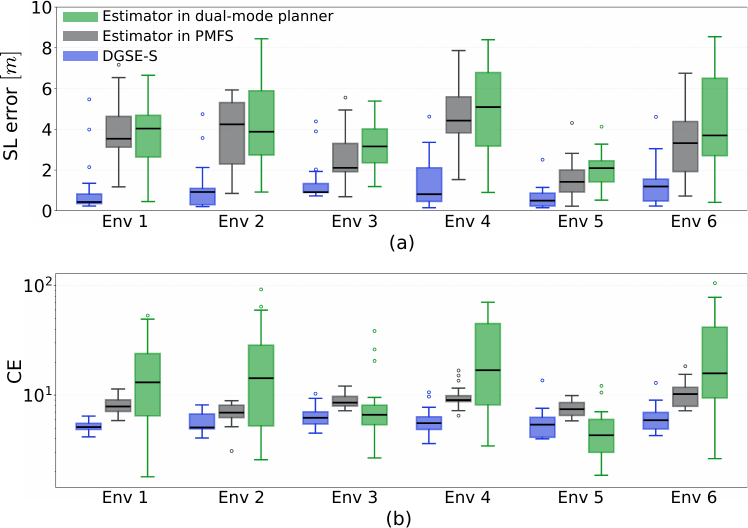}
    \vspace{-2mm}
    \caption{Source-localization performance from sparse random observations.
    SL error denotes source-localization error, and CE denotes the cross-entropy of the estimated source posterior.
    Each method was evaluated using 50 randomly sampled sensing locations over 20 trials in each environment.}
    \vspace{-3mm}
    \label{fig:gsl_only}
\end{figure}

We compared DGSE-S with the estimation components of representative indoor GSL methods that differ in how they construct the posterior distribution.
PMFS~\cite{ojeda2024robotic} estimates the source posterior using a filament-based gas dispersion model.
The dual-mode planner~\cite{kim2024gas} estimates the source posterior using a particle filter with an analytical dispersion model developed for negligible-airflow conditions.

For each environment, 50 sensing locations were randomly sampled from the free space.
Twenty independent trials were conducted per environment.
DGSE-S, PMFS, and the dual-mode planner were evaluated using the same sparse measurements, using only their source posterior estimation components and without executing their active GSL policies.
Performance was measured using the source-localization error and the CE of the estimated source posterior.
The source-localization error was computed from the probability-weighted source estimate defined in Section~\ref{sec3_active}.

As shown in Fig.~\ref{fig:gsl_only}, DGSE-S achieved lower source-localization errors than the comparison methods across all six environments.
DGSE-S also showed lower or comparable CE in most environments.

\subsection{Active GSL simulation in GADEN}\label{sec_e:indomain}
\begin{table*}[t]
\caption{Results of active GSL}
\label{tab:active_search_results}
\begin{center}
\begin{threeparttable}
\vspace{-6mm}
\setlength{\tabcolsep}{3pt}
\begin{tabular}{c c c c c c c}
\hline
\textbf{\textit{Environment}} & \textbf{\textit{Method}} & \textbf{\textit{SR [$\%$]}} & \textbf{\textit{Error [$m$]}} & \textbf{\textit{ST [$s$]}} & \textbf{\textit{TD [$m$]}} & \textbf{\textit{One-step Computation Time [$s$]}} \\
\hline
 & Dual-mode planner & 55 & 2.100 (1.717) & 437.84 (293.11) & 37.22 (18.80) & 0.331 (0.150) \\
1 & GrGSL & -- & -- & -- & -- & -- \\
inlet velocity: 0.5\,$\mathrm{m/s}$ & PMFS & 80 & 1.368 (1.404) & 535.59 (189.27) & 53.78 (26.79) & 7.795 (1.372) \\
release rate: 0.06\,$\mathrm{g/s}$ & DGSE-S (Ours) & \textbf{100} & \textbf{0.197} (0.076) & \textbf{65.69} (7.12) & \textbf{11.42} (0.54) & \textbf{0.108} (0.0105) \\
\hline
 & Dual-mode planner & 20 & 2.693 (1.536) & 366.83 (201.68) & 34.48 (16.07) & 0.360 (0.123) \\
2 & GrGSL & -- & -- & -- & -- & -- \\
inlet velocity: 1.0\,$\mathrm{m/s}$ & PMFS & 85 & 1.761 (1.853) & 1359.69 (471.74) & 140.45 (60.34) & 8.168 (1.632) \\
release rate: 0.03\,$\mathrm{g/s}$ & DGSE-S (Ours) & \textbf{100} & \textbf{0.769} (0.368) & \textbf{158.94} (29.26) & \textbf{23.73} (2.57) & \textbf{0.125} (0.0386) \\
\hline
 & Dual-mode planner & 45 & 2.956 (1.742) & 442.28 (269.29) & 49.02 (28.94) & 0.297 (0.164) \\
3 & GrGSL & -- & -- & -- & -- & -- \\
inlet velocity: 1.5\,$\mathrm{m/s}$ & PMFS & \textbf{100} & 1.259 (0.323) & 1124.84 (432.00) & 112.98 (50.58) & 8.713 (2.024) \\
release rate: 0.045\,$\mathrm{g/s}$ & DGSE-S (Ours) & \textbf{100} & \textbf{0.430} (0.118) & \textbf{107.52} (20.47) & \textbf{15.17} (1.89) & \textbf{0.117} (0.00871) \\
\hline
 & Dual-mode planner & 40 & 2.985 (2.165) & 312.82 (205.30) & 40.68 (26.36) & 0.313 (0.140) \\
4 & GrGSL & 0 & 7.367 (2.307) & 210.76 (117.24) & 12.85 (7.15) & 0.374 (0.181) \\
inlet velocity: 2.0\,$\mathrm{m/s}$ & PMFS & 80 & 1.175 (1.268) & 1747.89 (502.56) & 180.03 (62.42) & 8.606 (2.201) \\
release rate: 0.06\,$\mathrm{g/s}$ & DGSE-S (Ours) & \textbf{100} & \textbf{0.456} (0.428) & \textbf{136.27} (22.36) & \textbf{28.13} (2.54) & \textbf{0.114} (0.00688) \\
\hline
 & Dual-mode planner & 55 & 1.700 (0.962) & 234.61 (72.80) & 25.23 (9.01) & 0.508 (0.219) \\
5 & GrGSL & \textbf{100} & 0.678 (0.180) & 224.48 (60.55) & 11.21 (2.43) & 0.452 (0.146) \\
inlet velocity: 2.0\,$\mathrm{m/s}$ & PMFS & \textbf{100} & 0.709 (0.252) & 247.85 (42.08) & 18.91 (3.26) & 7.251 (1.216) \\
release rate: 0.06\,$\mathrm{g/s}$ & DGSE-S (Ours) & \textbf{100} & \textbf{0.319} (0.161) & \textbf{54.14} (21.96) & \textbf{9.78} (1.78) & \textbf{0.0929} (0.00604) \\
\hline
 & Dual-mode planner & 25 & 2.610 (0.886) & 224.25 (79.77) & 27.10 (8.32) & 0.384 (0.166) \\
6 & GrGSL & 95 & 0.750 (1.288) & 296.81 (47.20) & 17.08 (2.51) & 0.413 (0.218) \\
inlet velocity: 3.0\,$\mathrm{m/s}$ & PMFS & 90 & 0.688 (0.703) & 515.49 (132.21) & 41.99 (13.67) & 7.611 (1.668) \\
release rate: 0.09\,$\mathrm{g/s}$ & DGSE-S (Ours) & \textbf{100} & \textbf{0.376} (0.145) & \textbf{52.31} (13.08) & \textbf{14.98} (0.93) & \textbf{0.102} (0.00676) \\
\hline
\end{tabular}
\begin{tablenotes}
\footnotesize
\item[--] GrGSL results are not reported for Environments~1--3 because its publicly released implementation requires gas detection at the initial robot position to initiate localization.
\item[*] Values are reported as mean (standard deviation) over 20 trials per environment.
\item[] Bold values indicate the best result among the methods achieving the highest success rate in each environment.
\end{tablenotes}
\end{threeparttable}
\end{center}
\vspace{-5mm}
\end{table*}

\begin{figure}[t!]
    \centering
    \includegraphics[width=0.425\textwidth]{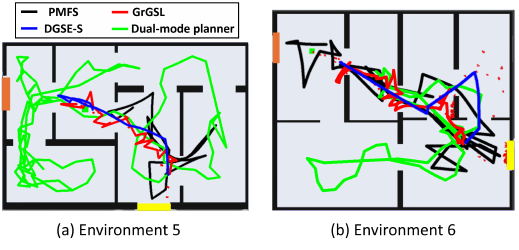}
    \vspace{-2mm}

    \caption{Search paths of DGSE-S and the comparison methods in Environments~5 and~6.}
    \vspace{-4mm}
    \label{fig:path_compare}
\end{figure}

\begin{figure}[t!]
    \centering
    \includegraphics[width=0.4\textwidth]{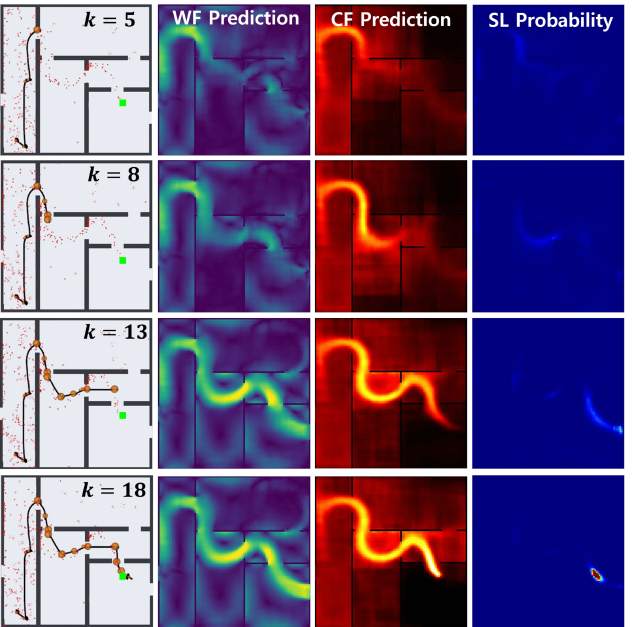}
    \vspace{-2mm}

    \caption{Illustrative results of DGSE-S source search in Environment~4.
    The black line represents the robot trajectory, while the orange circles indicate the locations where gas concentrations were measured.
    The circle sizes are proportional to the measured concentration.
    The panels show how the predicted wind field, concentration field, and source-location probability evolve over the search steps.
    The index $k$ denotes the step number, where each step corresponds to one sensing location visited by the robot.}
    \label{fig:dgse_search_process}
    \vspace{-2mm}
\end{figure}

We next evaluated active GSL in the same six GADEN environments.
This experiment assesses the complete online active GSL pipeline, including source-posterior estimation, goal selection, and termination.
DGSE-S was compared with PMFS~\cite{ojeda2024robotic}, the dual-mode planner~\cite{kim2024gas}, and GrGSL~\cite{ojeda2021information}.
GrGSL was excluded from the static source-inference experiment in Section~\ref{sec_e:gsl_only} because its posterior update is designed for sequential robot trajectories rather than independent sparse observations.
In active GSL, GrGSL was evaluated only in Environments~4--6 because its publicly released implementation requires gas detection at the initial robot position to initiate localization, which was not satisfied in Environments~1--3.

Each method was tested 20 times per environment and evaluated using five metrics: success rate (SR), source localization error, search time (ST), traveled distance (TD), and one-step computation time.
Source localization error, ST, and TD are computed over all trials, including failed trials; for timeout failures, the final estimate and elapsed values at the timeout are used.
For DGSE-S, the entropy-based termination parameter was set to $\eta=0.65$, as defined in Section~\ref{sec3_active}.
The one-step computation time refers to the total time of a single decision cycle, including input preprocessing, core inference, and utility or planner computation.
A trial was considered successful if the estimated source was within $2\,\mathrm{m}$ of the true source and the search was completed within $2{,}000\,\mathrm{s}$.
Following the hardware configuration described in Section~\ref{sec_e:gdm_comparison}, DGSE-S was executed on the Jetson AGX Orin, whereas the classical baselines were executed on the Intel NUC.

The results are summarized in Table~\ref{tab:active_search_results}.
DGSE-S achieved a 100\% SR in all six environments.
Among the methods achieving the highest SR in each environment, DGSE-S achieved the lowest source-localization error, shortest ST and TD, and lowest one-step computation time.
In Environment~4, GrGSL prematurely converged to incorrect source regions, likely due to its reliance on local upwind information, which was unreliable under the airflow pattern in this environment.
Overall, DGSE-S achieved more accurate and efficient active source search than the baseline methods across the evaluated environments.

The search paths of DGSE-S and the comparison methods in Environments~5 and~6 are shown in Fig.~\ref{fig:path_compare}.
Furthermore, to provide qualitative insight into the internal inference process, Fig.~\ref{fig:dgse_search_process} visualizes the source search process of DGSE-S in Environment~4.
The visualization shows the predicted wind field, concentration field, and source posterior over search steps.
An additional active GSL evaluation in out-of-domain environments is provided in Appendix~\ref{app:shift}.

\begin{figure}[t!]
    \centering                    
    \includegraphics[width=0.45\textwidth]{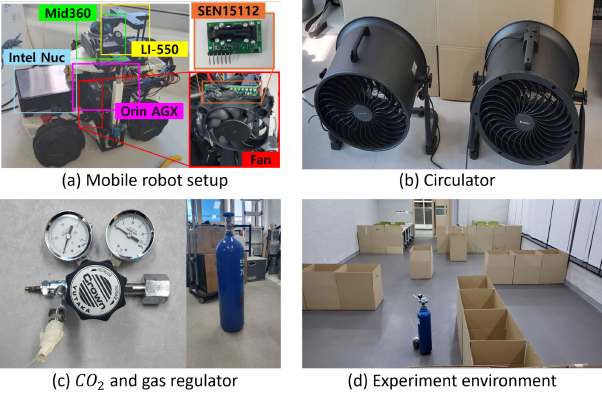}      
    \vspace{-4mm}      
    \caption{Setup for real-world experiments. The mobile robot carries LiDAR, gas, and wind sensors, while a fan establishes a structured indoor airflow and CO$_2$ is released as the gas source.}  \label{fig19} 
    \vspace{-3mm}                                                   
\end{figure}

\subsection{Real-world experiment for active GSL}\label{sec_e:real}

\begin{figure}
    \centering                   
    \includegraphics[width=0.425\textwidth]{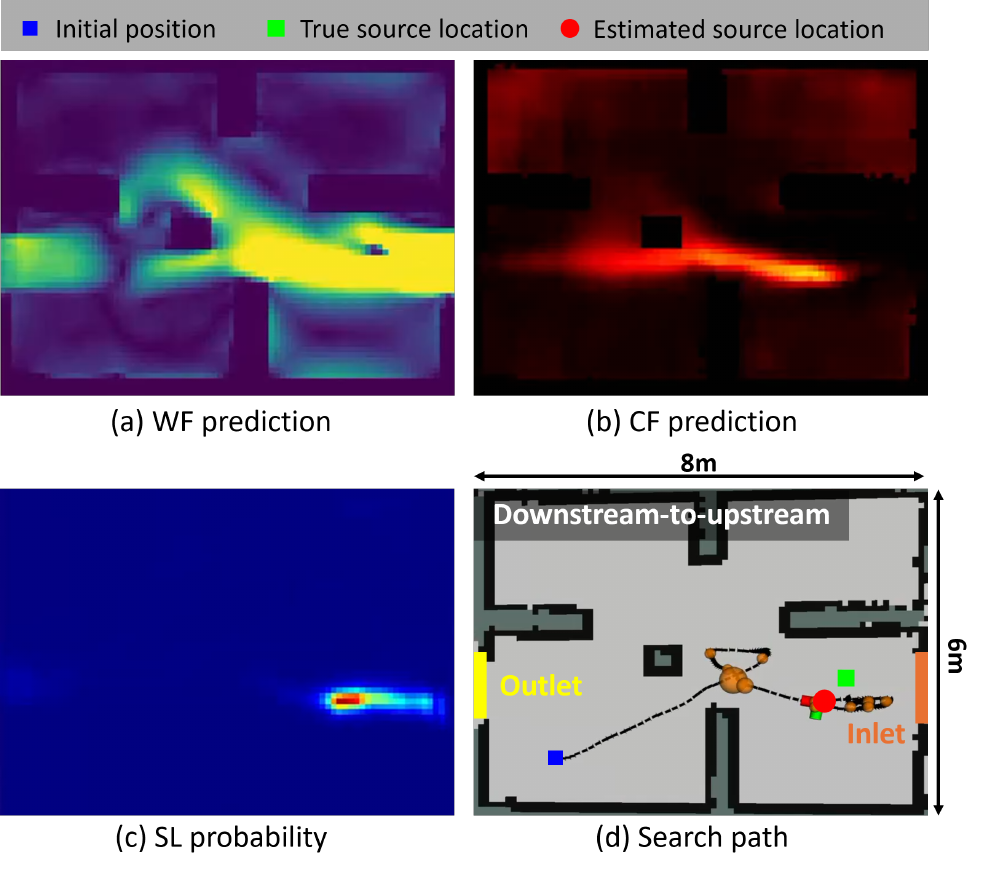}      
    \vspace{-3mm}   
    \caption{Illustrative results of real-world Experiment~1. The figure shows the robot trajectory, gas measurement locations, predicted concentration field, and source-location probability during downstream-to-upstream source search.}
    \vspace{-2mm}
    \label{fig20}                                                   
\end{figure}

\begin{figure}
    \centering                
    \includegraphics[width=0.425\textwidth]{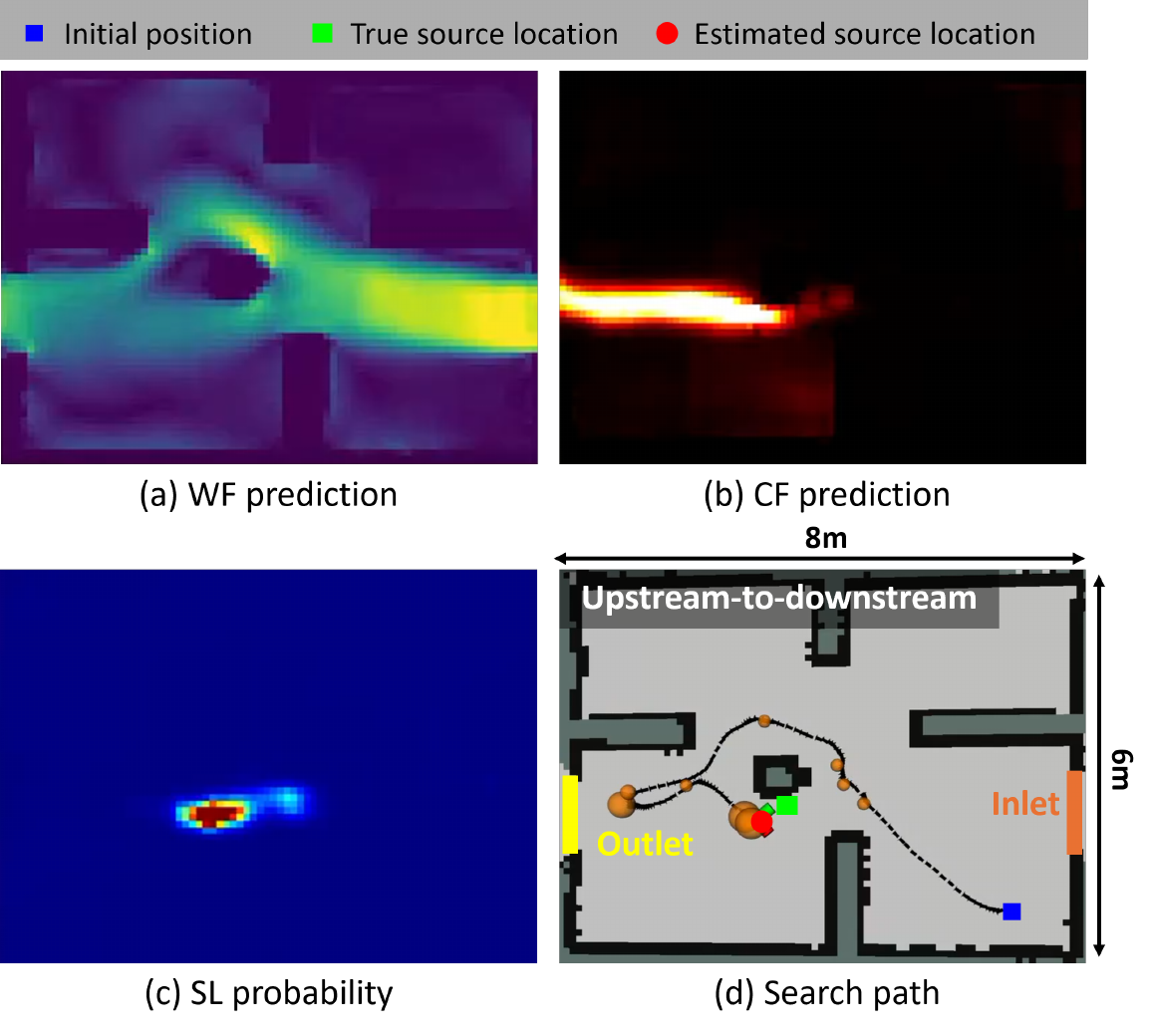}      
    \vspace{-3mm}   
    \caption{Illustrative results of real-world Experiment~2. The figure shows the robot trajectory, gas measurement locations, predicted concentration field, and source-location probability during upstream-to-downstream source search.}
    \vspace{-3mm}
    \label{fig21}                                                   
\end{figure}

To validate the proposed method beyond simulations, we conducted real-world active GSL experiments in an indoor laboratory environment.
The experimental platform consisted of an Agilex LIMO UGV equipped with a Livox MID-360 LiDAR for simultaneous localization and mapping, processed using FAST-LIO~\cite{xu2021fast}.
Gas concentration was measured using a SEN15112 module incorporating a nondispersive infrared (NDIR) CO$_2$ sensor, and wind speed and direction were measured using a TriSonica LI-550 anemometer.
The entire system was operated on ROS~1, with autonomous navigation handled by the \texttt{move\_base} navigation stack.
Consistent with the hardware setup described in Section~\ref{sec_e:gdm_comparison}, DGSE-S was executed on the Jetson AGX Orin onboard the robot, while the comparison methods were run on the Intel NUC.

CO$_2$ was used as the real-world gas source for safety reasons, while ethanol was used in simulation.
This substitution is reasonable for controlled validation because CO$_2$ and ethanol have similar molecular weights 
(44 vs.\ 46\,g/mol) and gas-phase diffusivities in air 
($\approx 1.6 \times 10^{-5}$ vs.\ $1.2 \times 10^{-5}$\,m$^2$/s), 
and both are heavier than air.
For network input, the measured CO$_2$ concentrations were converted using a fixed scale so that their input range was comparable to that of the simulated ethanol observations used for training.
CO$_2$ was released through a pressure regulator set to $0.075\,\text{MPa}$, and the fan generated a steady inlet flow consistent with the boundary-condition assumptions used in training.
Thus, these experiments are intended as controlled real-world validations under structured indoor airflow, rather than exhaustive tests across arbitrary real-world dispersion regimes.

\begin{table*}[t!]
\caption{{Results of active GSL in real-world experiments}}
\begin{center}
\setlength{\tabcolsep}{3pt} 
\begin{tabular}{c c c c c c }
\hline
 & \textbf{\textit{Method}}&\textbf{\textit{SR [$\%$]}}& \textbf{\textit{Error [$m$]}} & \textbf{\textit{ST [$s$]}}& \textbf{{ \textit{TD [$m$]}}}\\
\hline
 &Dual-mode planner & 40 & 2.353 (0.653) & 367.64 (69.47) & 19.31 (4.35) \\
Experiment 1 &PMFS & 0 & 1.620 (0.476) & 500.00 (0.00) & 54.00 (3.18)  \\
 &DGSE-S (Ours) & \textbf{100} & \textbf{0.818} (0.341) & \textbf{90.48} (38.78) & \textbf{11.97} (4.06)\\
\hline
 &Dual-mode planner & 60 & 1.293 (0.430) & 371.29 (169.47) & 25.57 (5.80) \\
Experiment 2 &PMFS & 60 & 1.024 (0.344) & 402.49 (109.55) & 55.89 (2.05)  \\
 &DGSE-S (Ours) & \textbf{100} & \textbf{0.459} (0.277) & \textbf{99.65} (19.92) & \textbf{19.21} (1.82)\\
\hline
\end{tabular}
\vspace{-3mm}
\label{exp_result}
\end{center}
\end{table*}

\begin{figure*}[t!]
    \centering                
    \includegraphics[width=0.75\textwidth]{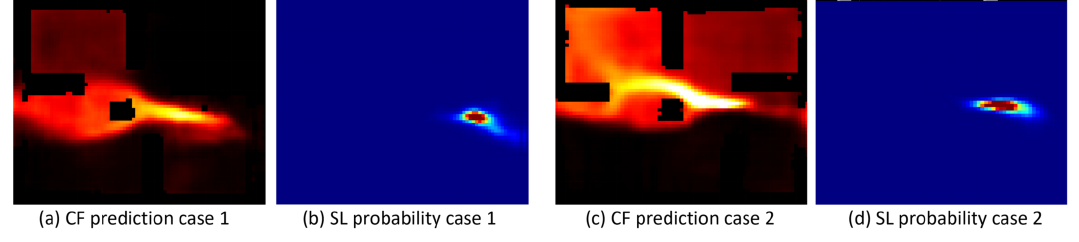}      
    \vspace{-3mm}   
    \caption{Concentration map and source probability map of DGSE-S in real-world Experiment~1, obtained via manual teleoperation. The maps illustrate how the continuous concentration estimate supports a source posterior despite secondary gas accumulation away from the primary transport stream.}
    \vspace{-3mm}
    \label{fig22}                                                   
\end{figure*}

\begin{figure*}[t!]
    \centering                
    \includegraphics[width=0.75\textwidth]{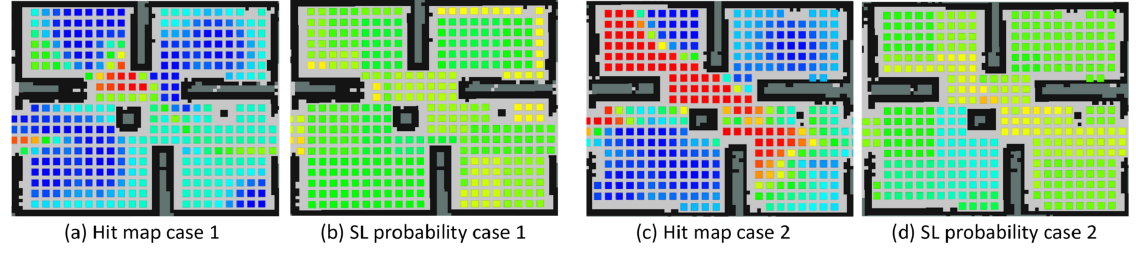}      
    \vspace{-3mm}   
    \caption{Gas hit map and source probability map of PMFS in real-world Experiment~1 for two representative trials. The binary hit maps show how thresholding can either miss the main transport stream or include secondary concentration regions.}
    \vspace{-3mm}
    \label{fig23}                                                   
\end{figure*}

Because gas detection at the initial robot position cannot be guaranteed in real deployments, the robot was initialized outside the gas transport region.
GrGSL requires gas detection at the initial position to trigger localization, as noted in Section~\ref{sec_e:indomain}, and was therefore excluded from the real-world comparison. 
The entropy-based termination parameter was set to $\eta=0.75$, and the utility weights were kept identical to those used in simulation.

Two contrasting scenarios were considered.
In Experiment~1, the robot started downstream of the source and had to navigate against the airflow to localize the upstream source.
In Experiment~2, the robot started upstream of the source and had to explore downstream after observing little or no gas signal.
Figure~\ref{fig19} shows the experimental setup, and Figs.~\ref{fig20}--\ref{fig21} illustrate representative results.

Each experiment was repeated 10 times.
A trial was considered successful if the declared source was within $2\,\text{m}$ of the true source and completed within $500\,\text{s}$.
The shorter timeout compared with simulation reflects practical constraints such as limited gas supply and robot battery capacity.
As summarized in Table~\ref{exp_result}, DGSE-S achieved a 100\% SR in both experiments and obtained lower localization error, shorter ST, and shorter TD than PMFS and the dual-mode planner.

\begin{figure}[t!]
    \centering                
    \includegraphics[width=0.375\textwidth]{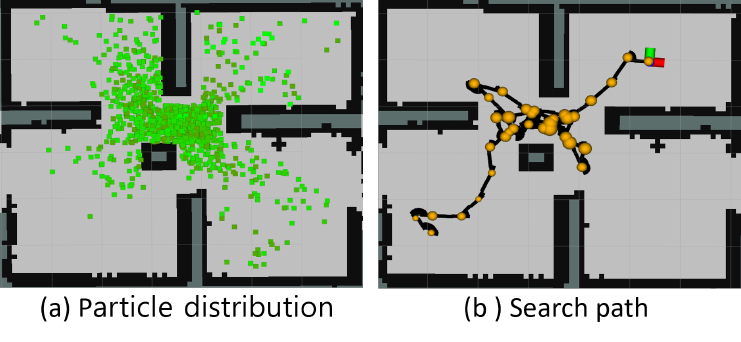}      
    \vspace{-3mm}   
    \caption{Illustrative results of the dual-mode planner in real-world Experiment~1. The path and concentration observations show convergence toward a high-concentration transport region rather than the true source under airflow-driven dispersion.}
    \vspace{-3mm}
    \label{fig24}                                                   
\end{figure}
To further understand these performance differences, we examine representative failure modes of the comparison methods in the real-world experiments.
A notable observation is the performance gap between DGSE-S and PMFS in Experiment~1.
As shown in Fig.~\ref{fig22}, part of the gas deviates from the primary transport stream and accumulates in the upper-left room.
PMFS estimates the source posterior from a binary gas hit map, which is obtained by applying a fixed concentration threshold to the gas measurements.
In this setting, such thresholding cannot reliably distinguish the primary plume from secondary concentration regions.
As illustrated by the two representative cases in Fig.~\ref{fig23}, the resulting hit map can either miss part of the main transport stream or include deviated concentration regions.
Consequently, PMFS failed to converge within the timeout in all Experiment~1 trials.

In contrast, DGSE-S infers the source location from continuous concentration fields rather than binarized measurements, making it more robust to deviations from the main transport stream.
In Experiment~2, where the gas is transported in a single coherent stream toward the outlet, PMFS performs comparatively better because the plume structure is more reliably captured by thresholding.
This is consistent with the in-domain results in Section~\ref{sec_e:indomain}, where PMFS performed better in Environments~5 and~6 with clearer single-stream gas transport.

The dual-mode planner underperformed mainly due to model mismatch: it assumes wind-free dispersion, whereas the real environment exhibits airflow-driven transport.
In Experiment~1, the long downstream path from the source to the outlet caused gas to remain concentrated over an extended transport region, leading the planner to converge toward high-concentration regions rather than the true source, as shown in Fig.~\ref{fig24}.
In Experiment~2, because the source-to-outlet transport path was shorter, even when the planner converged to a high-concentration region along the plume rather than the true source, the resulting localization error remained relatively small.

These results support the feasibility of online DGSE-S deployment for mobile-robot indoor GSL under the evaluated conditions.

\section{Conclusions and Future Work} \label{sec5}

\setcounter{table}{0}
\setcounter{figure}{0}
\setcounter{equation}{0}

\renewcommand{\thetable}{A.\arabic{table}}
\renewcommand{\thefigure}{A.\arabic{figure}}
\renewcommand{\theequation}{A.\arabic{equation}}
\renewcommand{\appendixprefix}{A.}

\begin{figure*}[t!]
    \centering                              
    \includegraphics[width=0.85\textwidth]{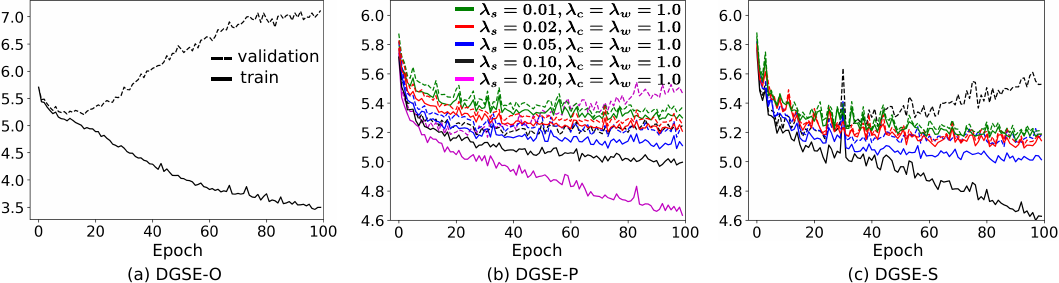}      
    \vspace{-3mm}   
    \caption{Source-localization (SL) loss over 100 epochs on the training and validation sets for DGSE-O, DGSE-P, and DGSE-S under different source-task weights.}  
    \vspace{-5mm}
    \label{fig4}                                                           
\end{figure*}

In this study, we proposed DGSE-S, a deep probabilistic method for indoor gas source localization in complex multi-room environments.
Its physical dependency-guided sequential inference explicitly incorporates the wind--concentration--source dependency structure underlying indoor gas dispersion, thereby enabling reliable source inference under sparse and noisy observations.

Comprehensive simulation and real-world experimental results demonstrated the effectiveness of DGSE-S for both source estimation and active GSL.
DGSE-S achieved the best overall concentration-field and source-localization performance among the DGSE variants.
The proposed model obtained lower concentration-field RMSE and NLL than those of representative indoor GDM baselines.
It also achieved lower source-localization errors than those of representative indoor GSL baselines.
In active GSL simulations, DGSE-S achieved strong localization accuracy and substantially improved search efficiency while supporting online embedded-GPU inference.
Finally, controlled real-robot experiments support the feasibility of online active GSL using DGSE-S under the evaluated real-world conditions.

The present study focuses on single-source localization under steady-state indoor dispersion, while broader deployment scenarios remain to be addressed.
In future work, we plan to extend the proposed framework toward broader deployment scenarios, including multi-source and transient dispersion settings, and to further investigate its integration with learning-based planning for improved search efficiency in large-scale environments.
We hope this work serves as a foundation for future research on deep learning--based gas source localization in realistic indoor scenarios.

\appendices
\section{Effect of Multi-Task Weighting on Training Behavior}\label{app:sloss}
To determine the training configuration used in the main experiments, we analyzed the effect of the source-task weight on the training behavior of the DGSE variants.
We compared DGSE-O, DGSE-P, and DGSE-S under the same training setup while varying the source-task weight for DGSE-P and DGSE-S.
DGSE-O, DGSE-P, and DGSE-S contain 0.82M, 1.54M, and 1.59M parameters, respectively, and require 1.61G, 3.15G, and 2.92G FLOPs for a single forward pass.
For DGSE-S, the reported FLOPs correspond to a single conditional forward pass.
DGSE-O is the most compact because it directly predicts only the source posterior.
All methods were trained on an NVIDIA RTX 3090 GPU using a batch size of 300, the Adam optimizer with a learning rate of $5\times10^{-4}$, and a weight decay of $1\times10^{-4}$.

The training and validation source-localization losses over 100 epochs are plotted in Fig.~\ref{fig4}.
As shown in Fig.~\ref{fig4}(a), DGSE-O exhibits unstable validation behavior despite its decreasing training loss.
This suggests that intermediate-field supervision can help stabilize source-posterior learning under sparse and noisy observations.
For DGSE-P and DGSE-S, the validation-loss behavior strongly depends on the relative weight assigned to the source task.
Large source-task weights tend to cause overfitting or instability, whereas excessively small weights lead to poor downstream source-localization performance.
Based on the validation source-localization loss and training stability, we set $\lambda_s=0.05$ and $\lambda_c=\lambda_w=1.0$ for both DGSE-P and DGSE-S in all main experiments.

\section{Validation of the Surrogate Objective}\label{app:val_object}

\setcounter{table}{0}
\setcounter{figure}{0}
\setcounter{equation}{0}

\renewcommand{\thetable}{B.\arabic{table}}
\renewcommand{\thefigure}{B.\arabic{figure}}
\renewcommand{\theequation}{B.\arabic{equation}}
\renewcommand{\appendixprefix}{B.}

\begin{table}[t!]
\caption{MC-estimated original losses on the validation set at the final epoch 
for varying $S_c$ and $S_s$}
\label{object}
\begin{center}
\setlength{\tabcolsep}{5pt}
\begin{tabular}{ccccccc}
\hline
$S_c = S_s$ & 1 & 2 & 5 & 10 & 50 \\
\hline
CF loss & $-$3.724 & $-$3.725 & $-$3.725 & $-$3.725 & $-$3.725 \\
SL loss & 5.153 & 5.104 & 5.090 & 5.092 & 5.091 \\
\hline
\end{tabular}
\end{center}
\vspace{-3mm}
\end{table}

\begin{figure}[t!]
    \centering                              
    \includegraphics[width=0.48\textwidth]{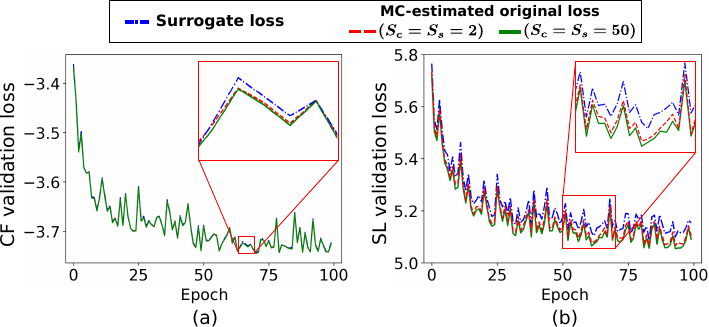}      
    \vspace{-3mm}   
    \caption{Comparison between the surrogate loss and the MC-estimated original losses over 2,000 validation environments (60,000 data points) during training, expressed in terms of losses.
(a) Concentration-field (CF) validation loss.
(b) Source-localization (SL) validation loss.} 
    \vspace{-3mm}
    \label{fig6}                                                           
\end{figure}

The DGSE-S training objective uses the Jensen's inequality--based surrogate objective in Eq.~\eqref{eq15}, rather than directly optimizing the original marginal log-likelihood. 
To examine whether this surrogate provides a practical training signal, we compare its training trend with Monte Carlo (MC) estimates of the original marginal objective. 
Since the original objective contains expectations inside logarithms and does not admit closed-form evaluation, we approximate the marginal log-likelihood terms as:
\begin{align}
    \log p(c_{\text{true}}|z^w,z^c,o) 
    \approx \log \left[\frac{1}{S_c} \sum_{i=1}^{S_c} 
    p_{\phi_c}(c_{\text{true}}|w_i,z^c,o)\right], \label{LBc}
\end{align}
\begin{align}
    \log p(s_{\text{true}}|z^w,z^c,o) 
    &\approx \log \Bigg[\frac{1}{S_c S_s} \sum_{j=1}^{S_s} 
    \sum_{i=1}^{S_c}  p_{\phi_s}(s_{\text{true}}|w_i,c_{ij},o)\Bigg], 
    \nonumber \\
    &\quad w_i \sim p_{\phi_w}, c_{ij} \sim p_{\phi_c}.
    \label{LBs}
\end{align}
Although this estimator is biased for finite samples, the bias decreases as the number of samples increases~\cite{burda2015importance}. 
For consistency with the training objective, Table~\ref{object} and Fig.~\ref{fig6} report the negative values of these objectives in loss form, such that lower values indicate better performance.
Table~\ref{object} reports the MC-estimated original loss at the 
final epoch for varying $S_c$ and $S_s$, evaluated over 60,000 validation samples. 
The loss changes negligibly once $S_c$ and $S_s$ reach 5, so we adopt $S_c=S_s=50$ as a numerically stable reference.
Figure~\ref{fig6}(a) compares the surrogate and MC-estimated original concentration-field (CF) losses during training, while Fig.~\ref{fig6}(b) shows the corresponding source-localization (SL) losses. 
For both tasks, the MC-estimated original losses ($S_c=S_s\in\{2,50\}$) exhibit training trends consistent with those of the surrogate objective throughout training, supporting the practical adequacy of the surrogate objective for training DGSE-S in the evaluated setting.

\section{Predictive Quality and Latency Trade-off in Sample-Based Inference}\label{app:sample}

\setcounter{table}{0}
\setcounter{figure}{0}
\setcounter{equation}{0}

\renewcommand{\thetable}{C.\arabic{table}}
\renewcommand{\thefigure}{C.\arabic{figure}}
\renewcommand{\theequation}{C.\arabic{equation}}
\renewcommand{\appendixprefix}{C.}

\begin{figure}[t!]
    \centering                              
    \includegraphics[width=0.45\textwidth]{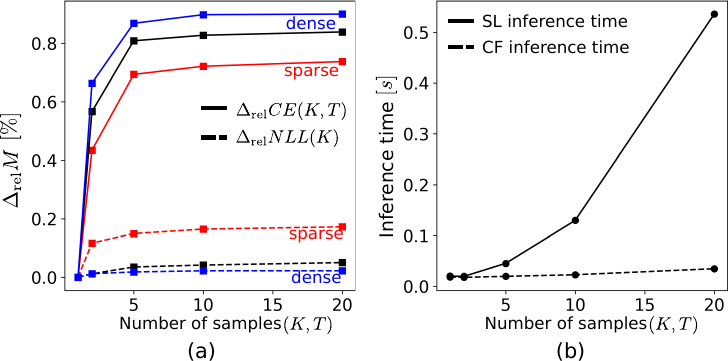}      
    \vspace{-3mm}   
    
    \caption{Effect of sample-based marginalization in DGSE-S.
(a) Relative reduction of source CE and concentration-field NLL compared with single-sample inference.
Black, red, and blue curves denote all validation samples, sparse-observation cases ($\leq 50$ observations), and dense-observation cases ($\geq 200$ observations), respectively.
(b) Inference latency of the concentration field (CF) and source-localization (SL) modules on an NVIDIA Jetson AGX Orin.}  \vspace{-1mm}
    \label{sample_compare}                                                           
\end{figure}

In DGSE-S, the concentration field distribution and source posterior are obtained by marginalizing over intermediate variables using sample-based inference, as described in Eqs.~\eqref{inf_c} and~\eqref{inf_s}.
The numbers of samples $K$ and $T$ therefore control a trade-off between predictive quality and inference latency.
To quantify the benefit of multiple samples, we report the relative reduction of the task-specific metric compared with single-sample inference:
\begin{equation}
\Delta_{\mathrm{rel}} M(K,T)
=
100 \times
\frac{M(1,1)-M(K,T)}{\vert M(1,1)\vert},
\end{equation}
where $M$ denotes either the CE of the source posterior or the NLL of the concentration field.
For the concentration field, the metric depends only on $K$, whereas the source CE depends on both $K$ and $T$ through the nested marginalization in Eq.~\eqref{inf_s}.

The evaluation was performed on the validation samples generated as described in Section~\ref{sec_data}, covering observation counts from 5 to 300.
In addition to the full validation set, we separately report sparse-observation cases with 50 or fewer observations and dense-observation cases with 200 or more observations.
As shown in Fig.~\ref{sample_compare}(a), increasing the number of samples improves both source CE and concentration-field NLL relative to single-sample inference.
The source CE improvement is larger in dense-observation cases, whereas the concentration-field NLL improvement is more pronounced in sparse-observation cases.

To assess online deployability, we measured inference latency on an NVIDIA Jetson AGX Orin.
As shown in Fig.~\ref{sample_compare}(b), the CF inference time remains nearly unchanged as $K$ increases within the tested range, because the additional samples can be efficiently processed through GPU batching.
In contrast, the SL inference time increases substantially at larger sample counts due to the nested sampling over both wind and concentration fields.
Based on these results, we set $K=T=5$ for DGSE-S in all main experiments, as this setting improves predictive quality over single-sample inference while maintaining low inference latency on the embedded GPU platform.

\section{Performance in Out-of-Domain Environments}
\label{app:shift}

\setcounter{table}{0}
\setcounter{figure}{0}
\setcounter{equation}{0}

\renewcommand{\thetable}{D.\arabic{table}}
\renewcommand{\thefigure}{D.\arabic{figure}}
\renewcommand{\theequation}{D.\arabic{equation}}
\renewcommand{\appendixprefix}{D.}

\begin{figure*}[t!]
    \centering                
    \includegraphics[width=0.85\textwidth]{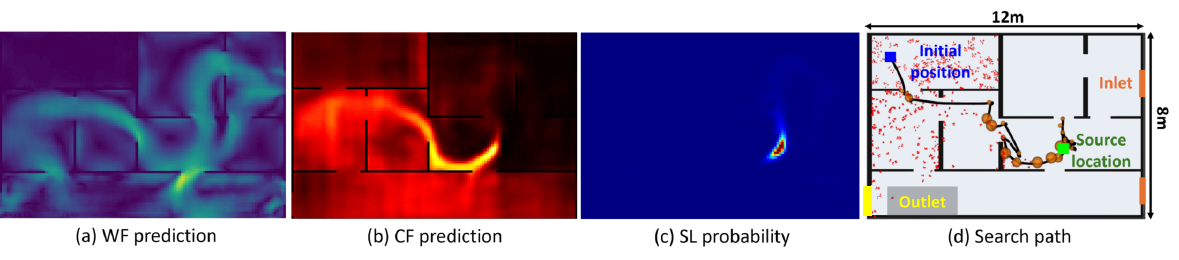}      
    \vspace{-3mm}   
    \caption{Illustrative DGSE-S active GSL results in OOD Environment~4.
The panels show the accumulated search path together with the predicted wind field, concentration field, and source posterior.}
    \vspace{-3mm}
    \label{fig18}                                                   
\end{figure*}

\begin{figure}[t]
    \centering                              
    \includegraphics[width=0.48\textwidth]{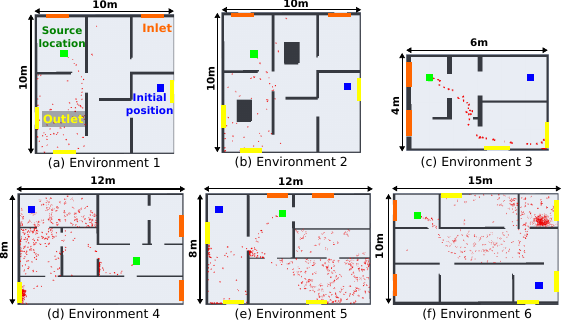}      
    \vspace{-3mm}   
    \caption{Out-of-domain environments used for the additional active GSL evaluation.
    The red dots denote simulated clusters of ethanol gas molecules.}
    \vspace{-3mm}
    \label{figout}                                                   
\end{figure}

\begin{table*}[t!]
\caption{Results of active GSL in out-of-domain environments.}
\vspace{-3mm}
\begin{center}
\setlength{\tabcolsep}{3pt} 
\begin{tabular}{c c c c c c c c c}
\hline
\textbf{\textit{Environment}} \textit{(inlet velocity, release rate)}&\textbf{\textit{SR [$\%$]}}& \textbf{\textit{Error [$m$]}} & \textbf{\textit{ST [$s$]}}& \textbf{{ \textit{TD [$m$]}}} & \textbf{{ \textit{One-step Computation Time [$s$]}}} & \textbf{{ \textit{PER}}}\\
\hline
1 (3.0\,$\mathrm{m/s}$, 0.03\,$\mathrm{g/s}$) & 100 & 0.851 (0.155) & 151.65 (30.76) & 27.82 (3.08) & 0.104 (0.0534) & 3.16 \\
2 (3.0\,$\mathrm{m/s}$, 0.03\,$\mathrm{g/s}$) & 100 & 0.881 (0.212) & 184.87 (76.20) & 18.39  (5.13) & 0.0966 (0.0444) &3.64 \\
3 (1.5\,$\mathrm{m/s}$, 0.09\,$\mathrm{g/s}$) & 100 & 0.540 (0.259) & 103.46 (20.53) & 14.41  (2.30) & 0.0858 (0.00239) &3.59 \\
4 (2.5\,$\mathrm{m/s}$, 0.06\,$\mathrm{g/s}$) & 100 & 0.593 (0.445) & 194.89 (76.65) & 26.94 (12.97) & 0.110 (0.0515) &3.00\\
5 (2.5\,$\mathrm{m/s}$, 0.09\,$\mathrm{g/s}$) & 100 & 0.952 (0.277) & 139.07 (85.02) & 21.36  (11.91) & 0.0943 (0.00307) &3.37\\
6 (1.5\,$\mathrm{m/s}$, 0.01\,$\mathrm{g/s}$) & 100 & 0.643 (0.391) & 374.78 (115.39) & 71.12 (17.63)& 0.135 (0.00621) &4.67\\
\hline
\end{tabular}
\begin{tablenotes}
\footnotesize
\item[*] PER denotes the path efficiency ratio, computed as TD divided by the shortest collision-free path length from the initial position to the true source.
\end{tablenotes}
\vspace{-6mm}
\label{de}
\end{center}
\end{table*}

Since DGSE-S is a learning-based method, an important practical question is whether it can still provide useful active GSL guidance in environments with configurations not represented during training.
To examine this question, we additionally evaluated DGSE-S in six out-of-domain (OOD) environments that differed from the training domain---partition-based multi-room layouts of 5--10 m with one inlet and one outlet---in terms of layout structure, obstacle configuration, physical dimensions, or inlet/outlet arrangement.
Figure~\ref{fig18} presents illustrative DGSE-S active GSL results in Environment 4, while Fig.~\ref{figout} shows the configurations of all six OOD environments.
Environment 1 contains two inlets and three outlets, Environment 2 additionally includes rectangular obstacles not represented in the training-layout generator, and Environments 3--6 combine multiple inlet/outlet configurations with physical dimensions outside the training range.
The $12\times8$ m and $15\times10$ m environments were represented using $120\times120$ and $150\times150$ grid canvases, respectively, while maintaining the training resolution of 0.1 m without spatial resampling or cropping.
Because the source posteriors were generally less concentrated in the OOD environments than in the in-domain environments, the termination threshold was increased from $\eta=0.65$ to $0.75$.

As summarized in Table~\ref{de}, DGSE-S achieved a 100\% success rate and a mean localization error below 1 m over 20 trials in each of the six OOD environments.
Search efficiency was evaluated using $\mathrm{PER}=\mathrm{TD}/d_{\mathrm{shortest}}$, where $d_{\mathrm{shortest}}$ denotes the shortest collision-free path length from the initial robot position to the ground-truth source location.
A PER value close to one therefore indicates that the robot reaches the source along a trajectory close to the shortest feasible path.
For reference, the PER values in the in-domain environments of Section~\ref{sec_e:indomain} were 1.18, 1.51, 1.77, 1.82, 2.89, and 2.17 for Environments 1--6, respectively, providing a mean of 1.89.
The mean PER increased to 3.57 in the OOD environments.
Thus, although DGSE-S retained successful source localization with low localization error outside the training domain, its search trajectories became less efficient.
Improving search efficiency in environments outside the training domain therefore remains an important direction for future work.

\bibliographystyle{IEEEtran}
\bibliography{reference.bib}

@article{Hutchinson17,
  title={{A review of source term estimation methods for atmospheric dispersion events using static or mobile sensors}},
  author={Hutchinson, Michael and Oh, Hyondong and Chen, Wen-Hua},
  journal={Inf. Fusion},
  volume={36},
  pages={130--148},
  year={2017},
  publisher={Elsevier}
}

@article{vergassola2007infotaxis,
  title={{'Infotaxis' as a strategy for searching without gradients}},
  author={Vergassola, Massimo and Villermaux, Emmanuel and Shraiman, Boris I},
  journal={Nature},
  volume={445},
  number={7126},
  pages={406--409},
  year={2007},
  publisher={Nature Publishing Group UK London}
}

@article{monroy2017gaden,
  title={{GADEN: A 3D gas dispersion simulator for mobile robot olfaction in realistic environments}},
  author={Monroy, Javier and Hernandez-Bennetts, Victor and Fan, Han and Lilienthal, Achim and Gonzalez-Jimenez, Javier},
  journal={Sensors},
  volume={17},
  number={7},
  pages={1479},
  year={2017},
  publisher={MDPI}
}

@inproceedings{prabowo2020bayesian,
  title={{A Bayesian approach for gas source localization in large indoor environments}},
  author={Prabowo, Yaqub Aris and Ranasinghe, Ravindra and Dissanayake, Gamini and Riyanto, Bambang and Yuliarto, Brian},
  booktitle={Proc. IEEE/RSJ Int. Conf. Intell. Robots Syst.},
  pages={4432--4437},
  year={2020}
}

@article{prabowo2023integration,
  title={{Integration of Bayesian inference and anemotaxis for robotics gas source localization in a large cluttered outdoor environment}},
  author={Prabowo, Yaqub A and Trilaksono, Bambang R and Hidayat, Egi MI and Yuliarto, Brian},
  journal={IEEE Access},
  volume={11},
  pages={22705--22713},
  year={2023},
  publisher={IEEE}
}

@article{ojeda2024robotic,
  title={{Robotic gas source localization with probabilistic mapping and online dispersion simulation}},
  author={Ojeda, Pepe and Monroy, Javier and Gonzalez-Jimenez, Javier},
  journal={IEEE Trans. Robot.},
  volume={40},
  pages={3551--3564},
  year={2024},
  publisher={IEEE}
}

@inproceedings{monroy2017online,
  title={{Online estimation of 2D wind maps for olfactory robots}},
  author={Monroy, Javier G and Jaimez, Mariano and Gonzalez-Jimenez, Javier},
  booktitle={Proc. ISOCS/IEEE Int. Symp. Olfaction Electron. Nose},
  pages={1--3},
  year={2017}
}

@inproceedings{jin2023towards,
  title={{Towards efficient gas leak detection in built environments: data-driven plume modeling for gas sensing robots}},
  author={Jin, Wanting and Rahbar, Faezeh and Ercolani, Chiara and Martinoli, Alcherio},
  booktitle={Proc. IEEE Int. Conf. Robot. Autom.},
  pages={7749--7755},
  year={2023}
}

@article{an2022receding,
  title={{Receding-horizon {RRT-Infotaxis} for autonomous source search in urban environments}},
  author={An, Seulbi and Park, Minkyu and Oh, Hyondong},
  journal={Aerosp. Sci. Technol.},
  volume={120},
  pages={107276},
  year={2022},
  publisher={Elsevier}
}

@article{hutchinson2019source,
  title={{Source term estimation of a hazardous airborne release using an unmanned aerial vehicle}},
  author={Hutchinson, Michael and Liu, Cunjia and Chen, Wen-Hua},
  journal={J. Field Robot.},
  volume={36},
  number={4},
  pages={797--817},
  year={2019},
  publisher={Wiley Online Library}
}

@article{hutchinson2018entrotaxis,
  title={{Entrotaxis as a strategy for autonomous search and source reconstruction in turbulent conditions}},
  author={Hutchinson, Michael and Oh, Hyondong and Chen, Wen-Hua},
  journal={Inf. Fusion},
  volume={42},
  pages={179--189},
  year={2018},
  publisher={Elsevier}
}

@article{hutchinson2018information,
  title={{Information-based search for an atmospheric release using a mobile robot: algorithm and experiments}},
  author={Hutchinson, Michael and Liu, Cunjia and Chen, Wen-Hua},
  journal={IEEE Trans. Control Syst. Technol.},
  volume={27},
  number={6},
  pages={2388--2402},
  year={2018},
  publisher={IEEE}
}

@article{park2020cooperative,
  title={{Cooperative information-driven source search and estimation for multiple agents}},
  author={Park, Minkyu and Oh, Hyondong},
  journal={Inf. Fusion},
  volume={54},
  pages={72--84},
  year={2020},
  publisher={Elsevier}
}

@article{park2022receding,
  title={{Receding horizon-based {Infotaxis} with random sampling for source search and estimation in complex environments}},
  author={Park, Minkyu and Ladosz, Pawel and Kim, Jongyun and Oh, Hyondong},
  journal={IEEE Trans. Aerosp. Electron. Syst.},
  volume={59},
  number={1},
  pages={591--609},
  year={2022},
  publisher={IEEE}
}

@inproceedings{endall2017uncertainties,
  title={What uncertainties do we need in bayesian deep learning for computer vision?},
  author={Kendall, Alex and Gal, Yarin},
  booktitle={Adv. Neural Inf. Process. Syst.},
  volume={30},
  pages={5574--5584},
  year={2017}
}

@inproceedings{kingma2013auto,
  title={{Auto-encoding variational {Bayes}}},
  author={Kingma, Diederik P and Welling, Max},
  booktitle={Proc. Int. Conf. Learn. Represent.},
  year={2014}
}

@inproceedings{burda2015importance,
  title={{Importance weighted autoencoders}},
  author={Burda, Yuri and Grosse, Roger and Salakhutdinov, Ruslan},
  booktitle={Proc. Int. Conf. Learn. Represent.},
  year={2016}
}

@article{farrell2003plume,
  title={{Plume mapping via hidden {Markov} methods}},
  author={Farrell, Jay A and Pang, Shuo and Li, Wei},
  journal={IEEE Trans. Syst., Man, Cybern., Part B (Cybern.)},
  volume={33},
  number={6},
  pages={850--863},
  year={2003},
  publisher={IEEE}
}

@article{kim2024gas,
  title={{Gas source localization in unknown indoor environments using dual-mode information-theoretic search}},
  author={Kim, Seunghwan and Seo, Jaemin and Jang, Hongro and Kim, Changseung and Kim, Murim and Pyo, Juhyun and Oh, Hyondong},
  journal={IEEE Robot. Autom. Lett.},
  volume={10},
  number={1},
  pages={588--595},
  year={2025},
  publisher={IEEE}
}

@article{he2024gas,
  title={{Gas source localization using dueling deep {Q}-network with an olfactory quadruped robot}},
  author={He, Yu and Cheng, Lei and Pan, YaDuo and Wang, DuanChu and Li, YuAo and Zheng, Han},
  journal={Int. J. Adv. Robot. Syst.},
  volume={21},
  number={3},
  note={{A}rt. no. 17298806241255797},
  year={2024},
  publisher={SAGE Publications}
}

@article{chen2021deep,
  title={{A deep {Q}-network for robotic odor/gas source localization: Modeling, measurement and comparative study}},
  author={Chen, Xinxing and Fu, Chenglong and Huang, Jian},
  journal={Measurement},
  volume={183},
  pages={109725},
  year={2021},
  publisher={Elsevier}
}

@article{lilienthal2004building,
  title={{Building gas concentration gridmaps with a mobile robot}},
  author={Lilienthal, Achim and Duckett, Tom},
  journal={Robot. Auton. Syst.},
  volume={48},
  number={1},
  pages={3--16},
  year={2004},
  publisher={Elsevier}
}

@inproceedings{lilienthal2009statistical,
  title={{A statistical approach to gas distribution modelling with mobile robots-the kernel DM+V algorithm}},
  author={Lilienthal, Achim J and Reggente, Matteo and Trincavelli, Marco and Blanco, Jose Luis and Gonzalez, Javier},
  booktitle={Proc. IEEE/RSJ Int. Conf. Intell. Robots Syst.},
  pages={570--576},
  year={2009}
}

@inproceedings{reggente2009using,
  title={{Using local wind information for gas distribution mapping in outdoor environments with a mobile robot}},
  author={Reggente, Matteo and Lilienthal, Achim J},
  booktitle={Proc. IEEE Sensors},
  pages={1715--1720},
  year={2009}
}

@article{g2016time,
  title={{Time-variant gas distribution mapping with obstacle information}},
  author={G. Monroy, Javier and Blanco, Jose-Luis and Gonzalez-Jimenez, Javier},
  journal={Auton. Robots},
  volume={40},
  pages={1--16},
  year={2016},
  publisher={Springer}
}

@article{gongora2020joint,
  title={{Joint estimation of gas and wind maps for fast-response applications}},
  author={Gongora, Andres and Monroy, Javier and Gonzalez-Jimenez, Javier},
  journal={Appl. Math. Model.},
  volume={87},
  pages={655--674},
  year={2020},
  publisher={Elsevier}
}

@inproceedings{winkler2022super,
  title={{Super-resolution for gas distribution mapping: convolutional encoder-decoder network}},
  author={Winkler, Nicolas P and Matsukura, Haruka and Neumann, Patrick P and Schaffernicht, Erik and Ishida, Hiroshi and Lilienthal, Achim J},
  booktitle={Proc. IEEE Int. Symp. Olfaction Electron. Nose},
  pages={1--3},
  year={2022}
}

@inproceedings{winkler2024gas,
  title={{Gas distribution mapping with radius-based, bi-directional graph neural networks (RABI-GNN)}},
  author={Winkler, Nicolas P and Neumann, Patrick P and Schaffernicht, Erik and Lilienthal, Achim J},
  booktitle={Proc. IEEE Int. Symp. Olfaction Electron. Nose},
  pages={1--3},
  year={2024}
}

@article{kamarudin2018integrating,
  title={{Integrating SLAM and gas distribution mapping (SLAM-GDM) for real-time gas source localization}},
  author={Kamarudin, Kamarulzaman and Md Shakaff, Ali Yeon and Bennetts, Victor Hernandez and Mamduh, Syed Muhammad and Zakaria, Ammar and Visvanathan, Retnam and Ali Yeon, Ahmad Shakaff and Kamarudin, Latifah Munirah},
  journal={Adv. Robot.},
  volume={32},
  number={17},
  pages={903--917},
  year={2018},
  publisher={Taylor \& Francis}
}

@article{visvanathan2020improved,
  title={{Improved mobile robot based gas distribution mapping through propagated distance transform for structured indoor environment}},
  author={Visvanathan, Retnam and Kamarudin, Kamarulzaman and Mamduh, Syed Muhammad and Toyoura, Masahiro and Ali Yeon, Ahmad Shakaff and Zakaria, Ammar and Kamarudin, Latifah Munirah and Mao, Xiaoyang and Abdul Shukor, Shazmin Aniza},
  journal={Adv. Robot.},
  volume={34},
  number={10},
  pages={637--647},
  year={2020},
  publisher={Taylor \& Francis}
}

@article{jasak2009openfoam,
  title={{OpenFOAM: Open source {CFD} in research and industry}},
  author={Jasak, Hrvoje},
  journal={Int. J. Nav. Archit. Ocean Eng.},
  volume={1},
  number={2},
  pages={89--94},
  year={2009},
  publisher={Elsevier}
}

@inproceedings{tian2025deep,
  title={{Deep learning based topography aware gas source localization with mobile robot}},
  author={Tian, Changhao and Wang, Annan and Fan, Han and Wiedemann, Thomas and Luo, Yifei and Yang, Le and Lin, Weisi and Lilienthal, Achim J and Chen, Xiaodong},
  booktitle={Proc. IEEE Int. Conf. Robot. Autom.},
  pages={4380--4386},
  year={2025}
}

@article{rhodes2023structurally,
  title={{Structurally aware 3D gas distribution mapping using belief propagation: A real-time algorithm for robotic deployment}},
  author={Rhodes, Callum and Liu, Cunjia and Chen, Wen-Hua},
  journal={IEEE Trans. Autom. Sci. Eng.},
  volume={21},
  pages={1623--1637},
  year={2023},
  publisher={IEEE}
}

@inproceedings{ronneberger2015u,
  title={{U-Net: Convolutional networks for biomedical image segmentation}},
  author={Ronneberger, Olaf and Fischer, Philipp and Brox, Thomas},
  booktitle={Proc. Int. Conf. Med. Image Comput. Comput.-Assist. Intervent.},
  volume={9351},
  pages={234--241},
  year={2015}
}

@article{ojeda2021information,
  title={{Information-driven gas source localization exploiting gas and wind local measurements for autonomous mobile robots}},
  author={Ojeda, Pepe and Monroy, Javier and Gonzalez-Jimenez, Javier},
  journal={IEEE Robot. Autom. Lett.},
  volume={6},
  number={2},
  pages={1320--1326},
  year={2021},
  publisher={IEEE}
}

@article{zhong2024awed,
  title={{AWED: Asymmetric wavelet encoder-decoder framework for simultaneous gas distribution mapping and gas source localization}},
  author={Zhong, Shutong and Zeng, Ming and Mao, Liang},
  journal={IEEE Trans. Instrum. Meas.},
  volume={73},
  pages={1--10},
  year={2024},
  publisher={IEEE}
}

@inproceedings{ruiz2024gas,
  title={{Gas source localization using physics-guided neural networks}},
  author={Ruiz, Victor Prieto and Hinsen, Patrick and Wiedemann, Thomas and Shutin, Dmitriy and Christof, Constantin},
  booktitle={Proc. IEEE Int. Symp. Olfaction Electron. Nose},
  pages={1--3},
  year={2024}
}

@article{xu2021fast,
  title={{{FAST-LIO}: A fast, robust {LiDAR}-inertial odometry package by tightly-coupled iterated extended {Kalman} filter}},
  author={Xu, Wei and Zhang, Fu},
  journal={IEEE Robot. Autom. Lett.},
  volume={6},
  number={2},
  pages={3317--3324},
  year={2021},
  publisher={IEEE}
}

@article{bourne2020decentralized,
  author  = {J. R. Bourne and M. N. Goodell and X. He and J. A. Steiner and K. K. Leang},
  title   = {Decentralized multi-agent information-theoretic control for target estimation and localization: Finding gas leaks},
  journal = {Int. J. Robot. Res.},
  volume  = {39},
  number  = {13},
  pages   = {1525--1548},
  year    = {2020}
}

@article{arain2021sniffing,
  author  = {M. A. Arain and V. H. Bennetts and E. Schaffernicht and A. J. Lilienthal},
  title   = {Sniffing out fugitive methane emissions: Autonomous remote gas inspection with a mobile robot},
  journal = {Int. J. Robot. Res.},
  volume  = {40},
  number  = {4},
  pages   = {782--814},
  year    = {2021}
}

@article{bourne2019coordinated,
  author  = {J. R. Bourne and E. R. Pardyjak and K. K. Leang},
  title   = {Coordinated {B}ayesian-based bioinspired plume source term estimation and source seeking for mobile robots},
  journal = {IEEE Trans. Robot.},
  volume  = {35},
  number  = {4},
  pages   = {967--986},
  year    = {2019}
}

@article{park2022source,
  title={Source term estimation using deep reinforcement learning with gaussian mixture model feature extraction for mobile sensors},
  author={Park, Minkyu and Ladosz, Pawel and Oh, Hyondong},
  journal={IEEE Robot. Autom. Lett.},
  volume={7},
  number={3},
  pages={8323--8330},
  year={2022},
}

@article{lee2025enhanced,
  title={Enhanced reward function Design for Source Term Estimation Based on deep reinforcement learning},
  author={Lee, Junhee and Jang, Hongro and Park, Minkyu and Oh, Hyondong},
  journal={IEEE Access},
  volume={13},
  pages={87777--87792},
  year={2025}
}

@article{wang2024exploration,
  title={An exploration-enhanced search algorithm for robot indoor source searching},
  author={Wang, Miao and Xin, Bin and Jing, Mengjie and Qu, Yun},
  journal={IEEE Trans. Robot.},
  volume={40},
  pages={4160--4178},
  year={2024}
}

@article{nam2026corrected,
  title={Corrected multi-fidelity surrogate model based on deep neural networks for predicting Aerodynamic loads of missile},
  author={Nam, Hansol and Yoo, Hanphil and Kim, Hyoungjin and Kim, Kyu Hong},
  journal={Int. J. Aeronaut. Space Sci.},
  volume={27},
  number={1},
  pages={190--206},
  year={2026},
}

\end{document}